\documentclass[12pt]{article}

\usepackage{amssymb}
\usepackage{amsthm,amsmath}
\usepackage{amsfonts,latexsym,graphicx}
\usepackage{eurosym}
\usepackage[dvips]{color}
\usepackage{url}
\usepackage{dsfont}   
\usepackage{natbib}
 \bibpunct{(}{)}{,}{a}{,}{,}

\DeclareSymbolFontAlphabet{\Bbb}{AMSb}
\newcommand{\fPL}{f_{L,\P,\lb}}

\newcommand{\fDDnln}{f_{L,\DD_n,\lb_n}}

\newcommand{\RLP}[1]{{{\cal R}_{L,\P}(#1)}}
\newcommand{\RLsP}[1]{{{\cal R}_{\Ls,\P}(#1)}}

\newcommand{\penf}{{\lb\hhnorm{f}}}
\newtheorem{Definition}{Definition}
\newtheorem{Theorem}[Definition]{Theorem}

\newtheorem{Proposition}[Definition]{Proposition}

\newenvironment{declaration}[1]{\trivlist \item[\hskip \labelsep{\em #1 }]\ignorespaces}{\endtrivlist}
\newenvironment{proofof}[1]{\begin{declaration}{#1.}}{\end{declaration}}

\newcommand{\bc}{\begin{center}}
\newcommand{\ec}{\end{center}}
\newcommand{\bi}{\begin{itemize}}
\newcommand{\ei}{\end{itemize}}
\newcommand{\be}{\begin{equation}}
\newcommand{\ee}{\end{equation}}
\newcommand{\beqna}{\begin{eqnarray*}}
\newcommand{\eeqna}{\end{eqnarray*}}
\newcommand{\bt}{\begin{tabular}}
\newcommand{\et}{\end{tabular}}\newcommand{\bnum}{\begin{enumerate}}
\newcommand{\enum}{\end{enumerate}}
\newcommand{\beq}{\begin{eqnarray}}
\newcommand{\eeq}{\end{eqnarray}}

\newcommand{\RP}[2]{{{\cal R}_{#1,\P}(#2)}}

\newcommand{\RPr}[2]{{\cal R}_{#1,{\P},\lambda}^{{reg}}(#2)}

\newcommand{\Ex}{\mathbb{E}}     
\newcommand{\N}{\mathds{N}}

\newcommand{\R}{\mathds{R}}

\newcommand{\qedr}{\hfill \quad $\square$}

\newcommand{\fP}{f_{L,\P,\lb}}

\newcommand{\Ls}{L^{\star}}

\newlength{\fixboxwidth}
\def \P {{\rm P}}    
\def \Q {{\rm Q}}    
\def \D {{\rm D}}

\def \lb        { \lambda }
\def \a         { \alpha }

\def \ve        { \varepsilon }  
\def \e         { \epsilon }     

\def \et        { \tilde{\eta}}

\newcommand{\snorm}[1] {\Vert #1 \Vert}

\newcommand{\ynorm}[2]{\mbox{$\left\Vert #1 \right\Vert_{{#2}}$}}

\newcommand{\hnorm}[1]{\left\Vert #1 \right\Vert_{\cH}}
\newcommand{\hhnorm}[1]{\left\Vert #1 \right\Vert_{\cH}^{2}}
\newcommand{\inorm}[1]{\left\Vert #1 \right\Vert_{\infty}}

\newcommand{\sLp}[1]{{\mathcal{L}_p(\mu)}}

\newcommand{\rem}[1]{}

\newcommand{\cH}{H}

\newcommand{\cF}{\mathcal{F}}
\newcommand{\diffb}[1]{\nabla^B_{#1}}

\newcommand{\BIF}{\mathrm{BIF}}

\newcommand{\cX}{\mathcal{X}}
\newcommand{\cY}{\mathcal{Y}}
\newcommand{\cXY}{\cX\times\cY}

\newcommand{\cXYR}{\cX\times\cY\times\R}
\newcommand{\cXX}{\cX\times\cX}
\newcommand{\cU}{U}

\newcommand{\DD}{\mathds{D}}
\newcommand{\PM}{\mathcal{M}_1}

\newcommand{\inP}    {\mbox{$\longrightarrow \hspace{-3.0ex} ^{{\rm {P}}}$}}

\newcommand{\RTnr}[2]{{\cal R}_{#1,{\D_n},\lambda}^{{reg}}(#2)}

\newcommand{\fDn}{f_{L,\D_n,\lb}}

\newcommand{\mcF}{\mathcal{F}}

\begin{document}


 \author{Andreas Christmann and Robert Hable \\ 
         Department of Mathematics \\
         University of Bayreuth
         }

\date{}

\title{Support Vector Machines for Additive Models:\\ Consistency and Robustness}



\maketitle

\begin{abstract}
  

\noindent
Support vector machines (SVMs) are special kernel based methods and belong to the most successful learning methods since more than a decade.
SVMs can informally be described as a kind of regularized M-estimators for functions and have demonstrated their usefulness in many complicated real-life problems. 
During the last years a great part of the statistical research on SVMs has concentrated on the question how to design SVMs such that they are universally consistent and statistically robust for nonparametric classification or nonparametric regression purposes. 
In many applications, some qualitative prior knowledge of the distribution $\P$ or of the unknown function $f$ to be estimated is present or the prediction function with a good interpretability is desired, such that a semiparametric model or an additive model is of interest. 

In this paper we mainly address the question how to design SVMs by choosing the 
reproducing kernel Hilbert space (RKHS) or its corresponding kernel to obtain consistent and statistically robust estimators in additive models. 
We give an explicit construction of kernels
--- and thus of their RKHSs --- which leads in combination with a Lipschitz continuous loss function to consistent and statistically robust SMVs for additive models.
Examples are quantile regression based on the pinball loss function, 
regression based on the $\e$-insensitive loss function, and
classification based on the hinge loss function.

\end{abstract}



\section{Introduction}
\noindent
Kernel methods such as  support vector machines belong to the most successful learning methods since more than a decade, see \citet{ScSm2002}.
Examples are classification or regression models where we have
an input space $\cX$, an output space $\cY$, some unknown probability
measure $\P$ on $\cXY$, and an unknown function $f:\cX\to\R$ which describes
the quantity of interest, e.g. the conditional quantile curve, of the conditional
distribution of $\P(\cdot|x)$, $x\in\cX$.
Support vector machines can informally be described as a kind of regularized M-estimators for functions and have demonstrated their usefulness in many complicated  high-dimensional  real-life problems.
Besides several other nice features, one key argument for using SVMs has been the so-called ``kernel trick'' \citep{SchoelkopfSmolaMueller1998a}, which decouples the 
SVM optimization problem from the domain of the samples, thus making it 
possible to use SVMs on virtually any input space $\cX$.
This flexibility is in strong contrast to more classical learning methods from both machine learning and non-parametric statistics, which almost always require input spaces $\cX\subset \R^d$.
As a result, kernel methods have been successfully used in various application 
areas that were previously infeasible for machine learning methods. As examples 
we refer to 
\emph{(i)} SVMs where using probability measures, e.g. histograms, as input samples, 
have been used to analyze histogram data and coloured images
\citep{HeinBousquet2005, SriperumbudurEtAl2009},
\emph{(ii)} SVMs for text classification and web mining \citep{Joachims2002, LaffertyLebanon2005}, and 
\emph{(iii)} SVMs with kernels from computational biology, e.g. kernels for trees and graphs \citep{SchoelkopfTsudaVert2004}.

For a data set $D_n=\big((x_1,y_1),\dots,(x_n,y_n)\big)$,
the \emph{empirical SVM} is defined as 
\be \label{SVMD}
  \fDn :=\arg\inf_{f\in\cH} \frac{1}{n}\sum_{i=1}^n L(x_i,y_i,f(x_i)) +\penf \,.
\ee
That is, SVMs are based on three key components: 
\emph{(i)} a convex \emph{loss function} $L:\cXYR\to[0,\infty)$ used to measure the quality of the prediction $f(x)$, 
\emph{(ii)} a \emph{reproducing kernel Hilbert space} (RKHS) $\cH$ of functions $f$ to specify the set of functions over which the expected loss is minimized, and 
\emph{(iii)} the \emph{regularization term} $\lb\hhnorm{f}$ to reduce the danger of overfitting and to guarantee the existence of a unique SVM even if $L$ is not strictly convex. 
The RKHS is often implicitly defined by specifying a \emph{kernel} 
$k: \cX\times\cX\to\R$. 
Details about the definition of SVMs and some examples will be given in 
Section \ref{sec:svm-setup}.

During the last years a great part of the statistical research on SVMs has concentrated on the central question how to choose the loss function $L$, the RKHS $H$ or its kernel $k$, and sequences of regularization parameters $\lb_n$ to guarantee that SVMs are universally consistent and statistically robust for classification and regression purposes.
In a nutshell, it turned out in a purely non-parametric setup that SVMs based on the combination of a Lipschitz continuous loss function and a bounded continuous kernel with a dense and separable RKHS are universally consistent with desirable
statistical robustness properties for \emph{any} probability measure $\P$ from which we observed the data set,
see, e.g., \citet{SteinwartChristmann2008a} and \citet{ChristmannVanMessemSteinwart2009} for details.
Examples are the combination of the Gaussian RBF-kernel with the pinball loss function for nonparametric quantile regression, with the $\e$-insensitive loss function for nonparametric regression, or with the hinge loss function for nonparametric classification, see Section \ref{sec:svm-setup}.  

Although a nonparametric approach is often the best choice in practice due
to the lack of prior knowledge on $\P$, a semiparametric approach or an 
additive model \citep{FriedmanStuetzle1981,HastieTibshirani1990} can also be valuable. 
For example, we may be interested due to practical reasons only in functions $f$ which offer a nice interpretation because an interpretable prediction function can be crucial if the prediction $f(x)$ has to be explainable to clients. This can be the case
if the prediction is the expected claim amount of a client and these predictions are the basis for the construction of an insurance tariff.   
Here we will mainly consider additive models although models with a multiplicative 
can also be of interest. 
More precisely, for some $s\in\N$, the input space $\cX$ is split up into 
$s\in\N$ non-empty spaces according to 
\be
  \cX = \cX_1 \times \ldots \cX_s
\ee
and only additive functions $f:\cX\rightarrow\R$ of the form
$$
  f(x_1,\dots,x_n)\;=\;f_1(x_1)+\dots+f_s(x_s)\,, \qquad x_j\in\cX_j\,,
$$
are considered, where $f_j:\cX_j\rightarrow\R$ for $j\in\{1,\dots,s\}$.

To our best knowledge, there are currently no results on consistency and statistical robustness published on SVMs based on kernels designed for additive models. 
Of course, one can use one of the purely nonparametric SVMs described above, but the hope is, that SVMs based on kernels especially designed for such situations may offer better results.

In this paper we address the question how to design specific SVMs for additive models. 
The main goal of this paper is that we give an explicit construction principle of kernels --- and thus of their RKHSs --- which leads in combination with a Lipschitz continuous loss function to consistent and statistically robust SMVs for additive models.
Examples are SVMs in additive models for quantile regression based on the pinball loss function, for regression based on the $\e$-insensitive loss function, and
for classification based on the hinge loss function.

The rest of the paper is organized as follows.
In Section \ref{sec:svm-setup} we collect some known results on loss functions, kernels
and their RKHSs, and on support vector machines. 
These results are needed to state our results on consistency and
statistical robustness of SVMs for additive models in Section \ref{sec:mainresult}.
Although we have so far no result on the rates of convergence, our numerical examples given in Section \ref{sec:examples} will demonstrate that SVMs based on kernels designed for additive models can easily outperform standard nonparametric SVMs if 
the assumption of an additive model is valid.
Section \ref{sec:discussion} contains the discussion.
All proofs are given in the Appendix.

\section{Background on support vector machines} \label{sec:svm-setup}

Let $\mathcal{X}$ be a complete separable metric space 
and let $\mathcal{Y}$ be a 
closed subset of $\mathds{R}$. We will
always use the respective Borel-$\sigma$-algebras. The set
of all probability measures on the Borel-$\sigma$-algebra
of $\cXY$ is denoted by 
$\mathcal{M}_{1}(\mathcal{X}\times\mathcal{Y})$\,.
The random input variables $X_1,\dots,X_n$ take their values in $\cX$
and the random output variables $Y_1,\dots,Y_n$ take their values in $\cY$. 
It is assumed that
$\,(X_{1},Y_{1}),\dots,(X_{n},Y_{n})\,$ are independent 
and identically distributed according to some unknown
probability measure
$\mathrm{P}\in\mathcal{M}_{1}(\mathcal{X}\times\mathcal{Y})$\,.
Since $\mathcal{Y}\subset\R$ is closed, $\P$ can be split
into the marginal distribution $\P_{\cX}$ on $\cX$ and the
conditional distribution $\P(\,\cdot\,|\,x\,)$ of $Y$ given $x$.

The goal is to find a good predictor $f:\cX\rightarrow\R$
which predicts the value $y$ of an output variable
after observing the value $x$ of the corresponding input variable.
The quality of a prediction $t=f(x)$
is measured by a \emph{loss function}
$$L\;:\;\;\cXY\times\R\rightarrow[0,\infty)\,,\qquad
  (x,y,t)\;\mapsto\;L(x,y,t)\;.
$$
It is assumed that $L$ is measurable and 
$L(x,y,y)=0$ for every $(x,y)\in\mathcal{X}\times\mathcal{Y}$
-- that is, the loss is zero if the prediction $t$ equals the
actual value $y$ of the output variable. 
In addition, we make the standard assumption that 
$$L(x,y,\cdot)\;:\;\;\mathds{R}\;\rightarrow\;[0,\infty)\,,\qquad
  t\;\mapsto\;L(x,y,t)
$$
is convex for every $(x,y)\in\mathcal{X}\times\mathcal{Y}$ and
that additionally the following uniform Lipschitz property is fulfilled
for some real number $|L|_{1}\in(0,\infty)$\,:
\begin{eqnarray}\label{def-uniformly-lipschitz}
  \sup_{(x,y)\in\mathcal{X}\times\mathcal{Y}}
  \big|L(x,y,t)-L(x,y,t^{\prime})\big|\;\;\leq\;\;
  |L|_{1}\cdot|t-t^{\prime}|
  \qquad\; \forall\,t,t^{\prime}\in\mathds{R}\;.
\end{eqnarray}
We restrict our attention to Lipschitz continuous loss functions
because the use of loss functions which are not Lipschitz continuous 
(such as the least squares loss 
which is only locally Lipschitz continuous on unbounded domains) 
usually conflicts with robustness; see, e.g., 
\citet[\S\,10.4]{SteinwartChristmann2008a}.

The quality of a (measurable) predictor 
$f:\mathcal{X}\rightarrow\mathds{R}$
is measured by the \emph{risk}
$$\mathcal{R}_{L,\mathrm{P}}(f)\;=\;
  \int_{\mathcal{X}\times\mathcal{Y}}L\big(x,y,f(x)\big)\,
  \mathrm{P}\big(d(x,y)\big)\;.
$$
By different choices of $\cY$ and the loss function $L$, different
purposes are covered by this setup -- e.g.\ binary
classification for $\cY=\{-1;+1\}$ and the \emph{hinge loss}
$$
  L_{\text{hinge}}(x,y,t):=\max\{0,1-yt\}\;,
$$ 
regression for $\cY=\R$ and the \emph{$\e$-insensitive loss} 
$$
 L_\e(x,y,t):=\max\{0, |y-t|-\e\}
$$ 
where $\e>0$, and
quantile regression for $\cY=\R$ and the \emph{pinball loss} 
\begin{equation}\label{loss:pinball-formel}
   L_\tau(x,y,t) := 
    \begin{cases}
    (\tau-1) (y-t), \quad \hfill {\textrm{~if~}} y-t < 0, \\
    \tau (y-t),  \hfill {\textrm{~if~}} y-t \ge 0, \\
    \end{cases}
\end{equation}
where $\tau\in(0,1)$.

An optimal predictor is a measurable function
$f^\ast:\mathcal{X}\rightarrow\mathds{R}$
which attains the minimal risk, called \emph{Bayes-risk}, 
$$\mathcal{R}_{L,\mathrm{P}}^\ast\;=\;
  \inf_{f:\mathcal{X}\rightarrow\mathds{R} \atop \text{measurable}}
  \mathcal{R}_{L,\mathrm{P}}(f)\;.
$$
The optimal predictor in a set $\mathcal{F}$ of
measurable functions
$f:\mathcal{X}\rightarrow\mathds{R}$
is an $f^\ast\in\mathcal{F}$
which attains the minimal risk 
$$\mathcal{R}_{L,\mathrm{P},\mathcal{F}}^\ast\;=\;
  \inf_{f\in\mathcal{F}}
  \mathcal{R}_{L,\mathrm{P}}(f)\;.
$$
For example, the goal of quantile regression is to estimate
a conditional quantile function,
i.e., a function $f^\ast_{\tau,\P}:\cX\rightarrow\R$ such that
$$\P\big((-\infty,f^\ast_{\tau,\P}(x)]\,\big|\,x\,\big)\,\geq\,\tau
  \qquad\text{and}\qquad
  \P\big([f^\ast_{\tau,\P}(x),\infty)\,\big|\,x\,\big)\,\geq\,1-\tau
$$
for the quantile $\tau\in(0,1)$.
If $f^\ast_{\tau,\P}\in\mcF$, then the conditional quantile function
$f^\ast_{\tau,\P}$ attains the minimal risk 
$\mathcal{R}_{L_{\tau},\mathrm{P},\mathcal{F}}^\ast$
for the pinball loss $L_{\tau}$ (with parameter $\tau$)
so that quantile regression can be done by trying to minimize
the risk $\mathcal{R}_{L_{\tau},\mathrm{P}}$ in $\mcF$.

One way to build a non-parametric predictor
$f$ is to use a support vector machine
\begin{equation}\label{fPlambda}
    \fP := \arg \inf_{f\in H} \RP{L}{f} + \lambda \hnorm{f}^2 \, ,
\end{equation}
where $H$ is a reproducing kernel Hilbert
space (RKHS) of a measurable \emph{kernel} $k: \cXX \to \R$, and
$\lambda>0$ is a regularization parameter to reduce the danger of
overfitting, see e.g., \citet{Vapnik1998},
\cite{ScSm2002} or \cite{SteinwartChristmann2008a}
for details.
The \emph{reproducing property} of $k$
states that, for all $f \in H$ and all $x \in \cX$,  
$$
  f(x)=\langle f,\Phi(x)\rangle_H
$$ 
where $\Phi:\cX\rightarrow H,\;x\mapsto k(\cdot,x)$ denotes
the \emph{canonical feature map}.
A kernel
$k$ is called \emph{bounded}, if 
$$
  \inorm{k} := \sup_{x \in \cX}\sqrt{k(x,x)}  < \infty\,.
$$
Using the reproducing property and
$\hnorm{\Phi(x)}=\sqrt{k(x,x)}$, we obtain the well-known
inequalities
\begin{equation}\label{wk1a}
  \inorm{f} \le \inorm{k} \hnorm{f} 
\end{equation}
and
\begin{equation}\label{wk1b}
  \inorm{\Phi(x)} \le \inorm{k} \hnorm{\Phi(x)} \leq \inorm{k}^2
\end{equation}
for all $f \in \cH$ and all $x\in\cX$.
As an example of a bounded kernel, we mention the popular \emph{Gaussian radial basis function}
(\emph{GRBF}) \emph{kernel} defined by 
\be \label{kernel:GRBF}
  k_{\gamma}(x,x')=\exp(-\gamma^{-2} \,\| x-x' \|^2_{\R^d}), ~~~x,x'\in\cX,
\ee
where $\gamma$ is some positive constant and $\cX\subset\R^d$. 
This kernel leads 
to a large RKHS which is dense in $L_1(\mu)$ for all 
probability measures $\mu$ on $\R^d$.
We will also consider the \emph{polynomial kernel}
$$k_{m,c}((x,x')=\big(\langle x,x'\rangle_{\R^d}+c\big)^m,
  ~~~x,x'\in\cX,
$$ 
where $m\in(0,\infty)$, $c\in(0,\infty)$ and $\cX\subset\R^d$. 
The \emph{dot kernel} is a special polynomial kernel with $c=0$ and $m=1$. 
The polynomial kernel is bounded if and only if $\cX$ is bounded.

Of course, the regularized risk 
$$
  \RPr{L}{f} := \RP{L}{f} + \lambda \hnorm{f}^2
$$ 
is in general not computable, because $\P$ is unknown.
However, the empirical distribution
$$
  \D_n=\frac{1}{n}\sum_{i=1}^n\delta_{(x_i,y_i)}
$$ 
corresponding to the data set $D_n=\big((x_1,y_1),\dots,(x_n,y_n)\big)$ 
can be used as an estimator of $\P$. Here
$\delta_{(x_i,y_i)}$ denotes the Dirac distribution in $(x_i,y_i)$.
If we replace $\P$ by $\D_n$ in {(\ref{fPlambda})}, we obtain the
regularized empirical risk $\RTnr{L}{f}$ and the empirical SVM $\fDn$.
Furthermore, we need analogous notions where $(x_i,y_i)$ is
replaced by random variables $(X_i,Y_i)$. Thus, we define
$$\mathds{D}_n\;=\;\frac{1}{n}\sum_{i=1}^n\delta_{(X_i,Y_i)}\;.
$$
Then, for every $\omega\in\Omega$, $\mathds{D}_n(\omega)$ is the
empirical distribution corresponding to the
data set 
$\big((X_1(\omega),Y_1(\omega)),\dots,(X_n(\omega),Y_n(\omega))\big)$
and, accordingly, 
$\mathcal{R}_{L,\mathds{D}_n,\lambda}^{reg}(f)$ denotes the mapping
$\Omega\rightarrow\R,\;\,\omega\mapsto
 \mathcal{R}_{L,\mathds{D}_n(\omega),\lambda}^{reg}(f)
$,
and 
$f_{L,\mathds{D}_n,\lambda}$ denotes the mapping
$\Omega\rightarrow H,\;\,\omega\mapsto
 f_{L,\mathds{D}_n(\omega),\lambda}
$\,.

Support vector machines $f_{L,\mathrm{P},\lambda}$ 
need not exist for every probability measure
$\mathrm{P}\in\mathcal{M}_{1}(\mathcal{X}\times\mathcal{Y})$;
for Lipschitz continuous loss functions it is sufficient
for the existence of $f_{L,\mathrm{P},\lambda}$
that $\int L(x,y,0)\,\P\big(d(x,y)\big)<\infty$.
This condition may be violated by heavy-tailed distributions $\P$
and, in this case, it is possible that $\RP{L}{f}=\infty$
for \emph{every} $f\in H$.

In order to enlarge the applicability
of support vector machines to heavy-tailed
distributions, the following extension has been developed
in \cite{ChristmannVanMessemSteinwart2009}.
Following an idea already used by \cite{Huber1967} for M-estimates in 
parametric models, a \emph{shifted loss function}
$L^{\ast}:\mathcal{X}\times\mathcal{Y}\times\mathds{R}
  \rightarrow\mathds{R}\,
$
is defined by
$$L^{\ast}(x,y,t)\;=\;L(x,y,t)-L(x,y,0)
  \qquad \forall\,(x,y,t)\in\mathcal{X}\times\mathcal{Y}\times\mathds{R}\;.
$$
Then, similar to the original loss function $L$, define
the $L^{\ast}$\,-\,risk by 
$$\RP{\Ls}{f}
     \;=\;\int L^{\ast}\big(x,y,f(x)\big)\,\mathrm{P}\big(d(x,y)\big)
$$ 
and the regularized $L^{\ast}$\,-\,risk by
$$\RPr{\Ls}{f}\;=\;
    \mathcal{R}_{L^{\ast},\mathrm{P}}(f)
    \,+\,\lambda\|f\|_{H}^2
$$
for every $f\in H$\,. In complete analogy to 
(\ref{fPlambda}), we define the support vector machine
based on the shifted loss function $L^{\ast}$ by
\begin{equation}\label{fPlambda-star}
    \fP := \arg \inf_{f\in H} \RP{\Ls}{f} + \lambda \hnorm{f}^2 \, .
\end{equation}
If the support vector machine $\fP$ defined by (\ref{fPlambda}) exists,
we have seemingly defined $\fP$ in two different 
ways now. However, the two definitions
coincide in this case and the 
following theorem summarizes some basic results of
\cite{ChristmannVanMessemSteinwart2009}.
\begin{Theorem}\label{theorem-L-star-trick-summary}
  Let $L$ be a convex and Lipschitz continuous loss function and
  let $k$ be a bounded kernel.
  Then, for every 
  $\mathrm{P}\in\mathcal{M}_{1}(\mathcal{X}\times\mathcal{Y})$
  and every $\lambda\in(0,\infty)$\,,
  there 
  exists a unique SVM $f_{L,\mathrm{P},\lambda}\in H$ which minimizes
    $\mathcal{R}_{L^{\ast},\mathrm{P},\lambda}^{reg}$\,, i.e.
    $$\mathcal{R}_{L^{\ast},\mathrm{P}}(f_{L,\mathrm{P},\lambda})
        \,+\,\lambda\|f_{L,\mathrm{P},\lambda}\|_{H}^2\;=\;
      \inf_{f\in H}\,\mathcal{R}_{L^{\ast},\mathrm{P}}(f)
          \,+\,\lambda\|f\|_{H}^2\;.
    $$
  If the support vector machine $f_{L,\mathrm{P},\lambda}$
  defined by (\ref{fPlambda}) exists,
  then the two definitions (\ref{fPlambda}) and 
 (\ref{fPlambda-star}) coincide.
\end{Theorem}

\section{Support vector machines for additive models} \label{sec:mainresult}
  \subsection{Model and assumptions}
              \label{sec-model-assumptions}

As described in the previous section, the goal is to minimize
the risk $f\mapsto\RLP{f}$ in a set $\mcF$ of 
functions $f:\cX\rightarrow\R$. In this article, we assume
an \emph{additive model}. Accordingly, let
$$\cX\;=\;\cX_{1}\times\dots\times\cX_s
$$
where $\cX_{1},\dots,\cX_s$ are non-empty sets.
For every $j\in\{1,\dots,s\}$, let $\mcF_j$ be a set of
functions $f_j:\cX_j\rightarrow\R$. Then, we only consider 
functions $f:\cX\rightarrow\R$ of the form
$$f(x_1,\dots,x_s)\;=\;
  f_1(x_1)+\dots+ f_s(x_s)
  \qquad\forall\,(x_1,\dots,x_s)\,\in\,\cX_{1}\times\dots\times\cX_s
$$
for $f_1\in\mcF_1,\dots,f_s\in\mcF_s$. Thus,
\begin{eqnarray}\label{def-cF}
  \cF\;:=\;
  \big\{f_1+\dots+f_s\,\,:\;
        \;f_j\in\mcF_j,\;\;1\leq j\leq s
  \big\}\;.
\end{eqnarray}
In (\ref{def-cF}), we have identified $f_j$ with the map
$\cX\rightarrow\R,\;(x_1,\dots,x_s)\mapsto f_j(x_j)$.

Such additive models can be treated by support vector machines
in a very natural way. For every $j\in\{1,\dots,s\}$,
choose a kernel $k_j$ on $\cX_j$ with RKHS $H_j$.
Then, the space of functions
$$H\;:=\;
  \big\{f_1+\dots+f_s\,\,:\;
        \;f_j\in H,\;\;1\leq j\leq s
  \big\}
$$
is an RKHS on $\cX=\cX_1\times\dots\times\cX_s$ with kernel
$k=k_1+\dots+k_s$; see Theorem 
\ref{theorem-sum-of-kernels-on-different-domains} below. 
In this way, SVMs can be used to fit additive models and SVMs
enjoy at least three appealing features:
First, it is guaranteed that the
predictor has the assumed additive structure 
$(x_1,\dots,x_s)\mapsto f_1(x_1)+\dots+f_s(x_s)$. Second,
it is possible to still use the standard SVM machinery 
including the kernel trick \citep[\S\,2]{ScSm2002}
and 
implementations
of SVMs -- just by selecting 
a kernel $k=k_1+\dots+k_s$. Third, the possibility to choose 
different kernels
$k_1,\dots,k_s$ offers a great flexibility. For example, take
$s=2$ and let $k_1$ be a GRBF kernel on $\R^{d_1}$ and
$k_2$ be a GRBF kernel on $\R^{d_2}$. Since the
RKHS of a Gaussian kernel is an infinite dimensional
function space, we get non-parametric estimates
of $f_1$ and $f_2$. As a second example, 
consider a semiparametric model with
$f=f_1+f_2$ where 
$f_1:x_1\mapsto f_1(x_1)$ is assumed to be a polynomial function
of order at most $m$
and $f_2:x_2\mapsto f_2(x_2)$ may be some complicated function.
Then, this semiparametric model can be treated by
simply taking a polynomial kernel on $\R^{d_1}$ for $k_1$ and
a GRBF kernel on $\R^{d_2}$ for $k_2$. This can be used,
for example, in order
to model changes in space (for $d_1\leq 3$ and $x_1$ 
specifying the location) or in time (for $d_1=1$ and
$x_1$ specifying the point in time).

\begin{Theorem}\label{theorem-sum-of-kernels-on-different-domains}
  For every $j\in\{1,\dots,s\}$, let $\cX_j$ be a non-empty set
  and
  $$k_j\;:\;\;\cX_j\times\cX_j\;\rightarrow\;\R\,,\qquad
    (x_j,x_j^\prime)\;\mapsto\;k_j(x_j,x_j^\prime)\;,
  $$
  be a kernel with corresponding RKHS $H_j$. Define
  $k=k_1+\dots+k_s$.
  That is,
  $$k\big((x_1,\dots,x_s),(x_1^\prime,\dots,x_s^\prime)\big) 
    \;=\;k_1(x_1,x_1^\prime)+\dots+k_s(x_s,x_s^\prime)
  $$
  for every $x_j,x_j^\prime\in\cX_j$, $j\in\{1,\dots,s\}$.
  Then, $k$ is a kernel on $\cX=\cX_1\times\dots\times\cX_s$
  with RKHS
  $$H\;:=\;
    \big\{f_1+\dots+f_s\,\,:\;
          \;f_j\in H,\;\;1\leq j\leq s
    \big\}
  $$
  and the norm of $H$, given in 
  (\ref{theorem-sum-of-kernels-on-different-domains-p2}), fulfills
  \begin{eqnarray}\label{theorem-sum-of-kernels-on-different-domains-1}
    \big\|f_1+\dots+f_s\big\|_H^2\;\leq\;
    \big\|f_1\big\|_{H_1}^2+\dots+\big\|f_s\big\|_{H_s}^2
    \quad\;\forall\,\,f_1\in H_1,\dots, f_s\in H_s\;.\quad
  \end{eqnarray}
\end{Theorem}

\bigskip

If not otherwise stated, we make the following assumptions
throughout the rest of the paper although
some of the results are also valid under more general conditions.

\medskip

\noindent
\textbf{Main assumptions}\label{assumptions}\hfill
\textit{
\begin{enumerate}
 \item[(i)]
  For every $j\in\{1,\dots,s\}$, the set $\cX_j$ is
  a complete, separable metric space; $k_j$ is a
  continuous and bounded kernel on $\cX_j$ with
  RKHS $H_j$. Furthermore, $k=k_1+\dots+k_s$ 
  denotes the kernel on $\cX=\cX_1\times\dots\times\cX_s$
  defined in Theorem
  \ref{theorem-sum-of-kernels-on-different-domains}
  and $H$ denotes its RKHS.
 \item[(ii)] The subset $\cY\subset\R$ is closed.
 \item[(iii)] The loss function $L$ is convex and 
  fulfills the uniform Lipschitz continuity 
  (\ref{def-uniformly-lipschitz}) with
  Lipschitz constant $|L|_1\in(0,\infty)$. In addition,
  $L(x,y,y)=0$ for every $(x,y)\in\cXY$.
\end{enumerate}
}
\noindent
Note that every closed subset of $\R^d$ is a complete,
separable metric space.  
We restrict ourselves to Lipschitz continuous loss functions
and continuous and bounded kernels because 
it has been shown earlier that these assumptions are
necessary in order to ensure good robustness properties; see e.g.\
\citet[\S\,10]{SteinwartChristmann2008a}. 
The condition $L(x,y,y)=0$ is quite natural and practically always
fulfilled -- it means that the loss of a correct prediction is 0.
Our assumptions cover many of the most interesting cases.
In particular, the hinge loss (classification),
the $\e$-insensitive loss (regression) and the pinball
loss (quantile regression) fulfill all assumptions. 
Many commonly used kernels are continuous. 
In addition, the Gaussian kernel is always bounded, the linear
kernel and all polynomial kernels are bounded
if and only if $\cX_j$ is bounded. From the assumption that the kernels
$k_j$ are continuous and bounded on $\cX_j$, it follows that
the kernel $k=k_1+\dots k_s$ is continuous and bounded on 
$\cX$.

  \subsection{Consistency} \label{sec:consistency} 
SVMs are called universally consistent, if the risk of the SVM estimator $\fDDnln$ converges, for \emph{all} probability measures $\P$, in probability to the Bayes-risk, i.e.
\be 
   \RLsP{\fDDnln} ~\inP~~ \mathcal{R}_{L^\ast,\P}^\ast \qquad (n\to\infty)\;.
\ee
In order to obtain universal consistency of SVMs,
it is necessary to choose a kernel with a large 
RKHS. Accordingly most known results about 
universal consistency of SVMs assume that the RKHS is dense in
$\mathcal{C}(\cX)$ where $\cX$ is a compact metric space
(see e.g.\ \cite{Steinwart2001a})
or, at least, that the RKHS is dense in $L_{q}(\P_{\cX})$
for some $q\in[1,\infty)$.
In this paper, we consider an additive model where
the goal is to minimize the risk
$f\mapsto\RLP{f}$ in the set
$$\cF\;=\;
  \big\{f_1+\dots+f_s\,\,:\;
        \;f_j\in\mcF_j,\;\;1\leq j\leq s
  \big\}\;.
$$
For the consistency of SVMs in an additive model, 
we do not need that the RKHS $H=H_1+\dots+H_s$ is dense
in the whole space $L_{q}(\P_{\cX})$; instead, we only assume that each
$H_j$ is dense in $\cF_j$.
As usual, $\mathcal{L}_q(\mu)$ denotes the set of all
$q$-integrable real-valued functions 
with respect to some measure $\mu$ and
$L_q(\mu)$ denotes the set of all equivalence classes in
$\mathcal{L}_q(\mu)$. Theorem \ref{theorem-consistency}
shows consistency of SVMs in additive models. That is,
the $L^\ast$-risk of $\fDDnln$ converges in probability to
the smallest possible risk in $\cF$.

\begin{Theorem}\label{theorem-consistency}
  Let the main assumptions (p.\ \pageref{assumptions}) be valid. Let
  $\P\in\mathcal{M}_1(\cXY)$ such that
  $$H_j\;\subset\;\cF_j\;\subset\;\mathcal{L}_1(\P_{\cX_j})\,,
    \qquad 1\leq j\leq s\,,
  $$
  and let $H_j$ be dense in $\cF_j$ with respect to
  $\|\cdot\|_{L_1(\P_{\cX_j})}$. Then, 
  for every sequence
  $(\lambda_n)_{n\in\N}\subset(0,\infty)$ such that
  $\lim_{n\rightarrow\infty}\lambda_n=0$ and
  $\lim_{n\rightarrow\infty}\lambda_n^2n=\infty$,
  $$\RP{\Ls}{\fDDnln}\;\longrightarrow\;\mathcal{R}_{\Ls,\P,\cF}^{\ast}
    \qquad\quad(n\rightarrow\infty)
  $$
  in probability.
\end{Theorem}

In general, it is not clear whether convergence of the risks
implies convergence of the SVM $\fDDnln$. However, the following
theorem will show such a convergence for quantile
regression in an additive model -- under the condition
that the quantile function $f^\ast_{\tau,\P}$ 
actually lies in 
$\cF=\cF_1+\dots+\cF_s$. In order to formulate this result,
we define
$$d_0(f,g,)\;=\;\int \min\big\{1\,,\,|f-g|\big\}\,d\P_{\cX}
$$
where $f,g:\cX\rightarrow\R$ are arbitrary measurable functions.
It is known that $d_0$ is a metric describing 
convergence in probability. 
\begin{Theorem}\label{theorem-quantile-regression-consistency}
  Let the main assumptions (p.\ \pageref{assumptions}) be valid. Let
  $\P\in\mathcal{M}_1(\cXY)$ such that
  $$H_j\;\subset\;\cF_j\;\subset\;\mathcal{L}_1(\P_{\cX_j})
    \qquad\forall\,j\in\{1,\dots,s\}
  $$
  and $H_j$ is dense in $\cF_j$ with respect to
  $\|\cdot\|_{L_1(\P_{\cX_j})}$. Let $\tau\in(0,1)$ and assume that
  the quantile function $f^\ast_{\tau,\P}$ is $\P_{\cX}$\,-\,almost
  surely unique and that
  $$f^\ast_{\tau,\P}\;\in\cF\;.
  $$
  Then, 
  for the pinball loss function $L=L_\tau$ and
  for every sequence
  $(\lambda_n)_{n\in\N}\subset(0,\infty)$ such that
  $\lim_{n\rightarrow\infty}\lambda_n=0$ and
  $\lim_{n\rightarrow\infty}\lambda_n^2n=\infty$,
  $$d_0\big(\fDDnln,f^\ast_{\tau,\P}\big)\;\longrightarrow\;0
    \qquad\quad(n\rightarrow\infty)
  $$
  in probability.
\end{Theorem}

  \subsection{Robustness} \label{sec:robustness} 
During the last years some general results on the statistical robustness properties of SVMs have been shown. Many of these results are directly applicable to SVMs for additive models if the kernel is bounded and continuous (or at least measurable)
and the loss function is Lipschitz continuous.
For brevity we only give upper bounds for the bias and the Bouligand
influence function for SVMs, which are both even applicable for non-smooth loss functions like the pinball loss for quantile regression,
and refer to \citet{ChristmannVanMessemSteinwart2009} and 
\citet[Chap.\,10]{SteinwartChristmann2008a} for results on the classical influence function proposed by \citep{Hampel1968,Hampel1974} and to
\citet{HableChristmann2009} for qualitative robustness of SVMs.

Define the function
\be \label{Tmap}
  T: \PM(\cXY)\to \cH, \quad T(\P) := \fPL \,,
\ee
which maps each probability distribution to its SVM.
In robust statistics we are interested in smooth and bounded functions $T$,
because this will give us stable SVMs within small
neighborhoods of $\P$. If an appropriately chosen derivative of $T(\P)$ is bounded, then we expect the value of $T(\Q)$ to be close to the value of $T(\P)$
for distributions $\Q$ in a small neighborhood of $\P$.

The next result shows that the $\cH$-norm of the difference of two SVMs
increases with respect to the mixture proportion $\ve\in(0,1)$ at most linearly in gross-error neighborhoods. 
The norm of total variation of a signed measure $\mu$ is denoted by $\snorm{\mu}_{\mathcal M}$.

\begin{Theorem}[Bounds for bias]\label{MR:bound-cor1}
If the main assumptions (p.\ \pageref{assumptions}) are valid,
then we have, for all $\lb>0$, all $\ve\in[0,1]$, and all probability
measures $\P$ and $\Q$ on $\cXY$, that
\begin{eqnarray}
 \label{biasbound1}
 \inorm{T(\Q) - T(\P)} 
 & \le & 
 c \,\snorm{\P-\Q}_{\mathcal M}\, ,\\
\label{biasbound2}
\inorm{T((1-\ve)\P+\ve\Q) - T(\P)} 
& \le &
c \,\snorm{\P-\Q}_{\mathcal M} \,\,\ve\, ,
\end{eqnarray}
where $c =  \frac{1}{\lb} \, \inorm{k}^{2} \, |L|_{1}$.
\end{Theorem}

Because of {(\ref{wk1b})}, there are analogous bias bounds of SVMs with respect to the
norm in $H$, if we replace $c$ by $\tilde{c}:=\frac{1}{\lb} \, \inorm{k} \, |L|_{1}$.

While F.R. Hampel's influence function is related to a G\^{a}teaux-derivative which is linear, the Bouligand influence function is related to the
Bouligand derivative which needs only to be \emph{positive homogeneous}. 
Because this weak derivative is less known in statistics, 
we like to recall its definition.
Let $E_1$ and $E_2$ be normed linear spaces.
A function $f: E_1 \to E_2$ is called \emph{positive homogeneous} if 
$f(\a x)=\a f(x)$ for all $\a \ge 0$ and for all $x \in E_1$.
If $U$ is an open subset of $E_1$, then a function $f: U \to E_2$ is
called \emph{Bouligand-differentiable} at a point $x_0 \in \cU$, 
if there exists a positive homogeneous function $\diffb{}f(x_0): \cU \to E_2$
such that 
\be \nonumber 
  \lim_{h \to 0} \frac{\ynorm{f(x_0+h)-f(x_0)-\diffb{}f(x_0)(h)}{E_2}}{\ynorm{h}{E_1}} = 0,
\ee
see \citet{Robinson1991}.

The \emph{Bouligand influence function} (BIF) of the map $T:\PM(\cX\times\cY) \to \cH$ for a distribution $\P$ in the direction of a distribution $\Q \ne \P$ was defined by \citet{ChristmannVanMessem2008} as 
\begin{equation}
  \lim_{\ve \downarrow 0}
  \frac{\hnorm{T\bigl((1\!-\!\varepsilon)\P\!+\!\varepsilon\Q\bigr)\!-\!T(\P) -
        \mathrm{BIF}(\Q; T,\P)}}{\ve}  = 0 \label{BIF1}.
\end{equation}
Note that the BIF is a special Bouligand-derivative
\begin{equation*}
  \lim_{\snorm{\varepsilon(\Q-\P)} \to 0}
  \frac{\hnorm{T\bigl((\P + \varepsilon(\Q-\P)\bigr) - T(\P) -
        \mathrm{BIF}(\Q; T,\P)}}{\snorm{\varepsilon(\Q-\P)}}  = 0 
\end{equation*}
due to the fact, that $\Q$ and $\P$ are fixed, 
and it is independent of the norm on $\PM(\cXY)$.
The partial Bouligand derivative with respect to the third argument of $\Ls$
is denoted by $\diffb{3}\Ls(x,y,t)$.
The BIF shares with F.R. Hampel's influence function the interpretation that it measures the impact of an infinitesimal small amount of contamination of the original distribution $\P$ in the direction of $\Q$ on the quantity of
interest $T(\P)$. It is thus desirable that the function $T$ has a
\emph{bounded} BIF.
It is known that existence of the BIF implies existence of the IF and in this case 
they are equal.
The next result shows that, under some conditions, the Bouligand influence function of SVMs exists and is bounded, see \citet{ChristmannVanMessemSteinwart2009} for more related results.

\begin{Theorem}[Bouligand influence function]\label{MRTh1}
Let the main assumptions (p.\ \pageref{assumptions}) be valid, but assume that $\cX$ is a complete separable normed linear space.\footnote{Bouligand derivatives are only defined in normed linear spaces. E.g., $\cX\subset\R^d$ a linear subspace.} Let $\P,\Q\in\PM(\cX\times\cY)$.
Let $L$ be the pinball loss function $L_\tau$ with $\tau\in(0,1)$ or let 
$L$ be the $\e$-insensitive loss function $L_\e$ with $\e>0$. 
Assume that for all $\delta>0$ there exist positive constants
$\xi_\P$, $\xi_\Q$, $c_\P$, and $c_\Q$ such that for all $t\in\R$ with
$|t-\fP(x)| \le \delta \inorm{k}$ the following inequalities hold for all
      $a\in[0,2\delta\inorm{k}]$ and $x\in \cX$:
\be \label{assumption2a}
        \P\bigl( [t, t+a]\,\big|\,x\bigr) \le c_\P a^{1+\xi_\P} {~and~}
        \Q\bigl( [t, t+a]\,\big|\,x\bigr) \le c_\Q a^{1+\xi_\Q} \, .
\ee
Then the Bouligand influence function $\mathrm{BIF}(\Q;T,\P)$ of $T(\P):=\fPL$ exists, is bounded, and equals
\be \label{MRBIF2b}
\frac{1}{2} 
\Bigl( \Ex_{\P}\diffb{3}\Ls(X,Y,\fPL(X)) \, \Phi(X)  
- 
       \Ex_{\Q}\diffb{3}\Ls(X,Y,\fPL(X)) \, \Phi(X) \Bigr)\,. 
\ee
\end{Theorem}

Note that the Bouligand influence function of the SVM only depends on $\Q$ via the second term in {(\ref{MRBIF2b})}.
The interpretation of the condition {(\ref{assumption2a})} is that
the probability that $Y$ given $x$ is in some small interval around the 
SVM is essentially at most proportional to the length of the interval 
to some power greater than one. 

For the pinball loss function, the BIF given in {(\ref{MRBIF2b})} simplifies to
\be 
  \begin{split} \label{MRBIF2c}
  & \frac{1}{2\lambda}  
   \int_\cX \bigl(\P\bigl( (-\infty,\fPL(x)]\,\big|\,x\bigr) - \tau \bigr) \Phi(x) \, \P_\cX(dx) \\
  - &
  \frac{1}{2\lambda}  
  \int_\cX \bigl( \Q\bigl( (-\infty, \fPL(x)]\,\big|\,x\bigr) - \tau \bigr) \Phi(x) \, \Q_\cX(dx) .
\end{split}
\ee
The BIF of the SVM based on the pinball loss function can hence be interpreted as the difference of the integrated and with $\frac{1}{2\lb}\,\Phi(x)$ weighted difference between the estimated quantile level and the desired quantile level $\tau$.

Recall that the BIF is a special Bouligand derivative and thus positive homogeneous in $h=\ve (\Q-\P)$. 
If the BIF exists, we then immediately obtain 
\beq
  f_{L,(1-\alpha \ve)\P+\alpha\,\ve\Q, \,\lb} - \fP   & = &   T(\P + \alpha h) - T(\P) \nonumber \\
   & = &  \alpha \, \BIF(\Q; T, \P) + o(\alpha h) \label{poshom} \\
   & = & \alpha \bigl( T(\P + h) - T(\P)  +  o(h) \bigr) + o(\alpha h) \nonumber\\
   & = &  \alpha \bigl( f_{L,(1-\ve)\P+\ve\Q, \,\lb} - \fP \bigr)
          + o(\alpha\ve(\Q-\P)) \nonumber
\eeq 
for all $\alpha \ge 0$.
This equation gives us a nice approximation of the asymptotic bias term $f_{L,(1-\ve)\P+\ve\Q, \,\lb}-\fP$, if we consider the amount $\alpha \ve$ of contamination instead of $\ve$.

\section{Examples} \label{sec:examples} 
In this section we would like to illustrate our theoretical results on SVMs for additive models with a few finite sample examples.
The goals of this short section are twofold.
We like to get some preliminary
insight how SVMs based on kernels designed for additive models work 
for finite sample sizes when compared to the standard 
GRBF kernel defined on the whole input space  
and to get some ideas for further research on this topic. 
We also like to apply support vector machines based on the additive kernels
treated in this paper to a real-life data set.

\subsection{Simulated example}
Let us consider the following situation of median regression.
We have two independent input variables $X_1$ and $X_2$ each with a uniform distribution on the interval $[0, 1]$ and the output variable $Y$ given $x=(x_1,x_2)$ has a Cauchy distribution (and thus not even the first moment does exist) with center $f(x_1,x_2):=f_1(x_1)+f_2(x_2)$, where $f_1(x_1):=7+5 x_1^2$ and $f_2(x_2):=\sin(5 x_2) \cos(17 x_2)$. 
Hence the true function $f$ we like to estimate with SVMs has an additive
structure, where the first function is a polynomial of order two and
the second function is a smooth and bounded function but no polynomial.
Please note, that here $\cX=[0,1]^2$ is bounded whereas $\cY=\R$ is
unbounded. As $\cX$ is bounded, even a polynomial kernel on $\cX$ is bounded
which is not true for unbounded input spaces.
We simulated three data sets of this type with sample sizes $n=500$, 
$n=2,\!000$, and $n=10,\!000$. 
We compare the exact function $f$ with three SVMs 
$f_{L,\D,\lb_n}$ fitted by the three data sets, where we use the pinball loss function with $\tau=0.5$ because we are interested in median regression.
\bi
\item \emph{Nonparametric SVM}. 
We use an SVM based on the 2-dimensional GRBF kernel $k$ defined in {(\ref{kernel:GRBF})} to fit $f$ in a totally nonparametric manner.
\item \emph{Nonparametric additive SVM}. 
We use an SVM based on the kernel $k=k_1+k_2$ where $k_1$ and $k_2$ are 1-dimensional GRBF kernels.
\item \emph{Semiparametric additive SVM.} 
We use an SVM based on the kernel $k=k_1+k_2$ where $k_1$ is a polynomial kernel of order 2 to fit the function $f_1$ and $k_2$ is a 1-dimensional GRBF kernel to fit the function $f_2$.
\ei
Our interest in these examples is to check how well SVMs using kernels designed for additive models perform in these situations.
No attempt was made to find optimal values of the regularization parameters $\lb$ and the kernel parameter $\gamma$ by using a grid search or cross-validation, because we did not want to mix the quality of such optimization strategies with the choice of the kernels. We therefore fixed $\gamma=2$ and used the simple non-stochastic specification $\lb_n=0.05 n^{-0.45}$ for the regularization parameter which guarantees that our consistency result from Section \ref{sec:mainresult} is applicable.

\begin{figure}[b]
\caption{Quantile regression using SVMs and pinball loss function with $\tau=0.5$. Model: $Y|(x_1, x_2) \sim f_1(x_1)+f_2(x_2)+\mbox{Cauchy-errors}$, where $f_1(x_1):=7+5 x_1^2$ and $f_2(x_2)=\sin(5 x_2) \cos(17 x_2)$ and $x_1$ and $x_2$ are observations of independent and identically uniform distributed random variables on the interval $[0, 1]$. The regularization parameter is $\lb_n=0.05 n^{-0.45}$, and the kernel parameter of the Gaussian RBF kernel is $\gamma=2$. 
Upper left subplot: true function $f(x_1,x_2)=f_1(x_1)+f_2(x_2)$.
Upper right subplot: SVM fit based on GRBF kernel $k$ on $\cX=\R^2$.
Lower left subplot: SVM fit based on the sum of two 1-dimensional GRBF kernels.
Lower right subplot: SVM fit based on the sum of a 1-dimensional polynomial 
kernel on $\R$ and a 1-dimensional GRBF kernel.
\label{FigureN10000}\newline}
\begin{tabular}{ccc}
 \multicolumn{3}{c}{$\fbox{$n=10,\!000$}$} \\
 true function & $~~~~~~~~~~$ & nonparametric SVM \\[1.5cm]

\hspace{-1.2cm}
\includegraphics[width=4.3cm,angle=0]{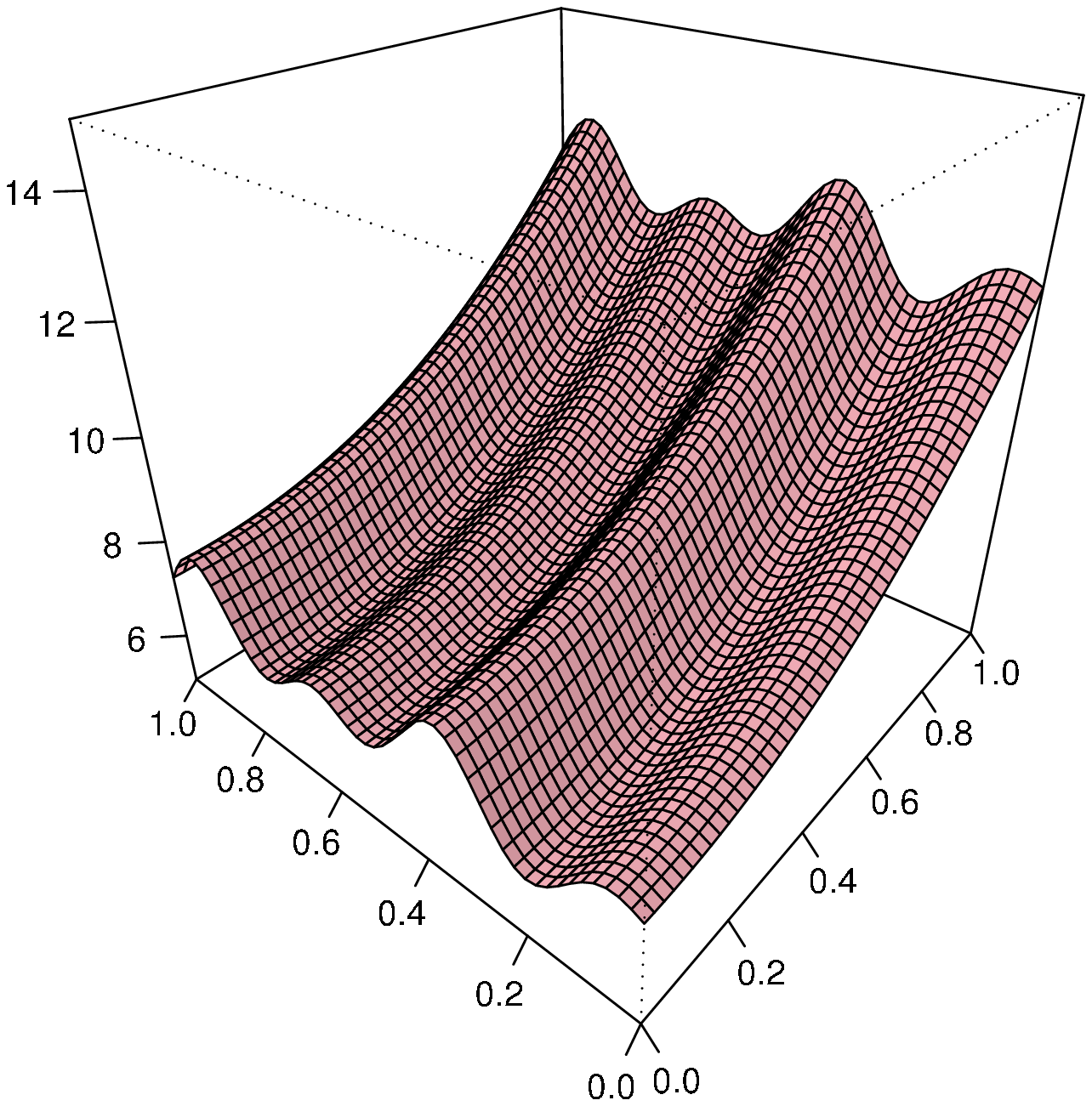} & &
\hspace{-2.cm}
\includegraphics[height=4.6cm,angle=0]{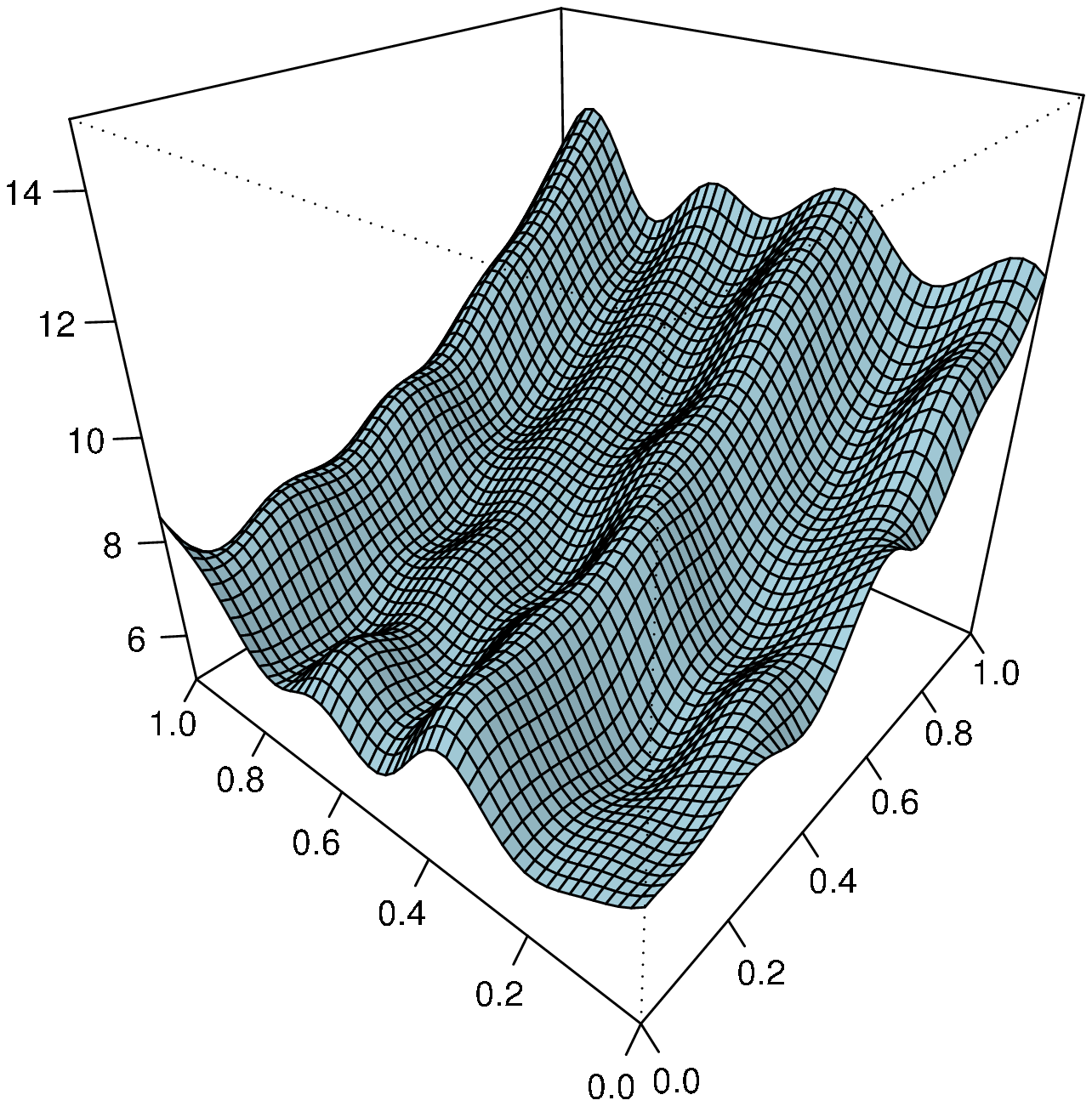} \\[1.cm]

 nonparametric additive SVM & &  semiparametric additive SVM \\[1.5cm]
\hspace{-1.2cm}
\includegraphics[width=4.6cm, angle=0]{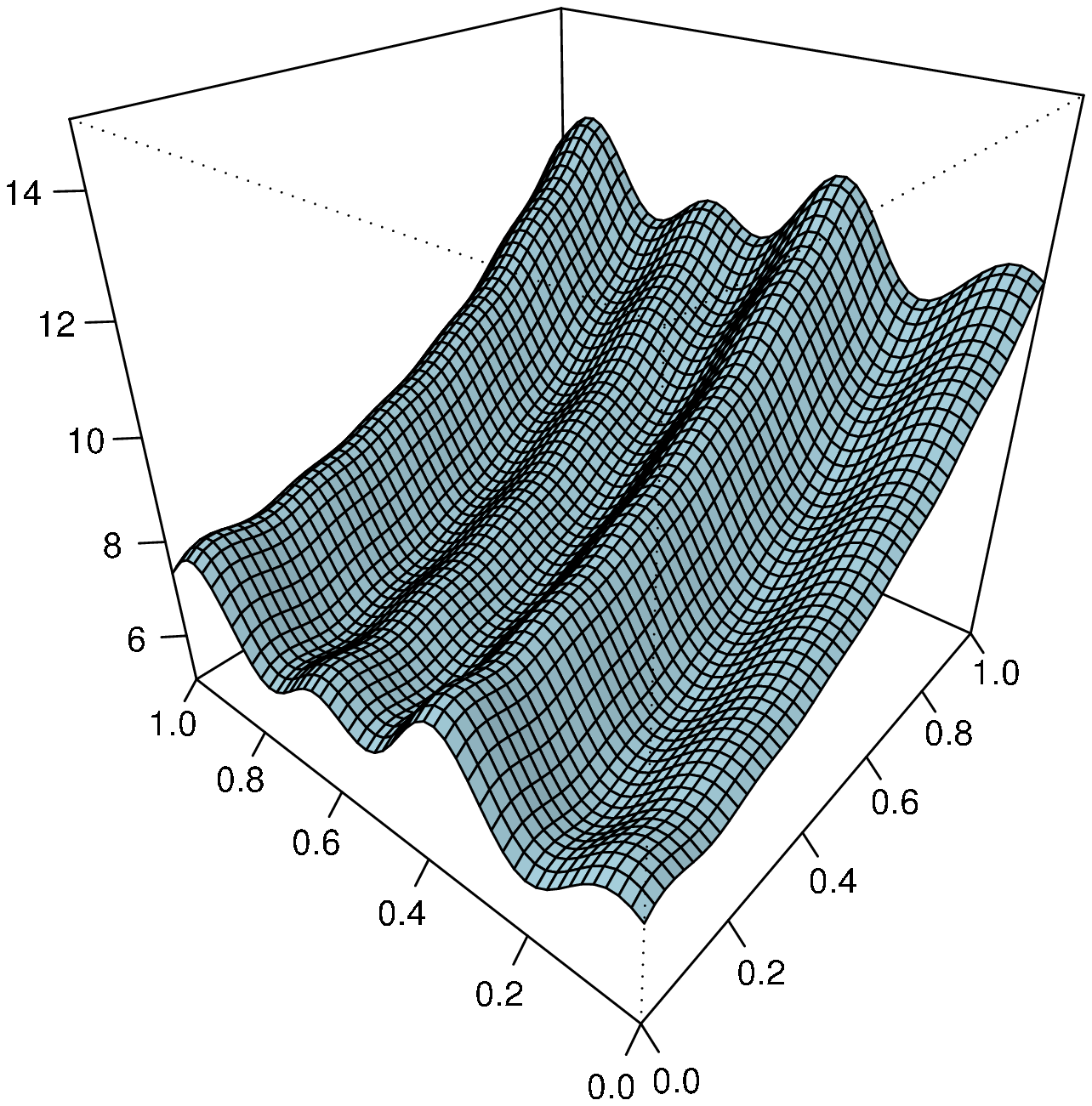} & &
\hspace{-2.cm}
\includegraphics[width=4.6cm, angle=0]{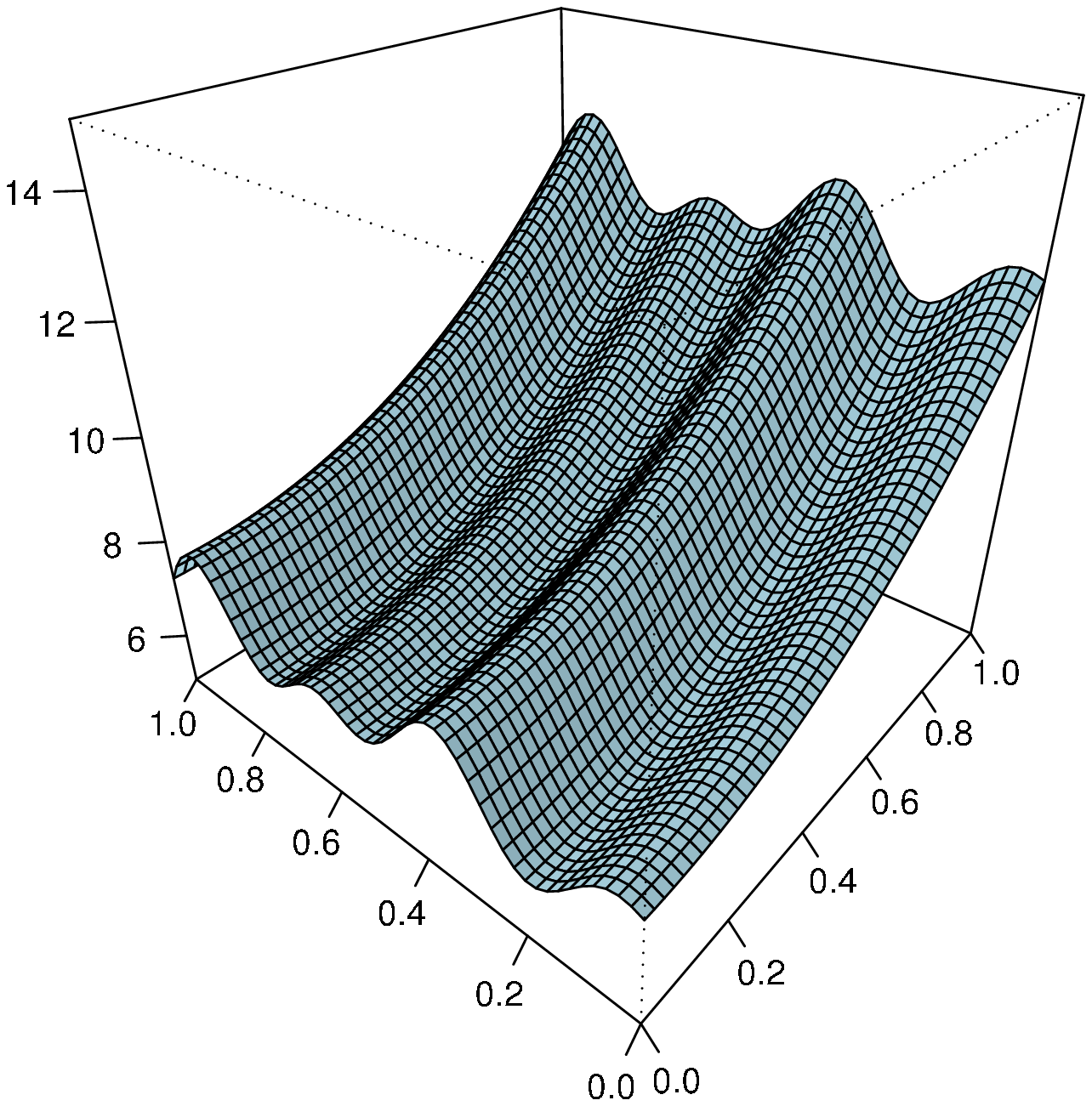} 
\end{tabular}
\end{figure}

\begin{figure}[t]
\caption{Quantile regression using SVMs and pinball loss function with $\tau=0.5$. Model: $Y|(x_1, x_2) \sim f_1(x_1)+f_2(x_2)+\mbox{Cauchy-errors}$, where $f_1(x_1):=7+5 x_1^2$ and $f_2(x_2)=\sin(5 x_2) \cos(17 x_2)$ and $x_1$ and $x_2$ are observations of independent and identically uniform distributed random variables on the interval $[0, 1]$. The regularization parameter is $\lb_n=0.05 n^{-0.45}$, and the kernel parameter of the Gaussian RBF kernel is $\gamma=2$. Upper left subplot: true function $f(x_1,x_2)=f_1(x_1)+f_2(x_2)$.
Upper right subplot: SVM fit based on GRBF kernel $k$ on $\cX=\R^2$.
Lower left subplot: SVM fit based on the sum of two 1-dimensional GRBF kernels.
Lower right subplot: SVM fit based on the sum of a 1-dimensional polynomial 
kernel on $\R$ and a 1-dimensional GRBF kernel.\label{FigureN2000}\newline}
\begin{tabular}{ccc}
 \multicolumn{3}{c}{$\fbox{$n=2,\!000$}$} \\
 true function & $~~~~~~~~~~$ & nonparametric SVM \\[1.5cm]
\hspace{-1.2cm}  
\includegraphics[width=4.3cm,angle=0]{exact3d.eps} & &
\hspace{-2.0cm}
\includegraphics[height=4.6cm,angle=0]{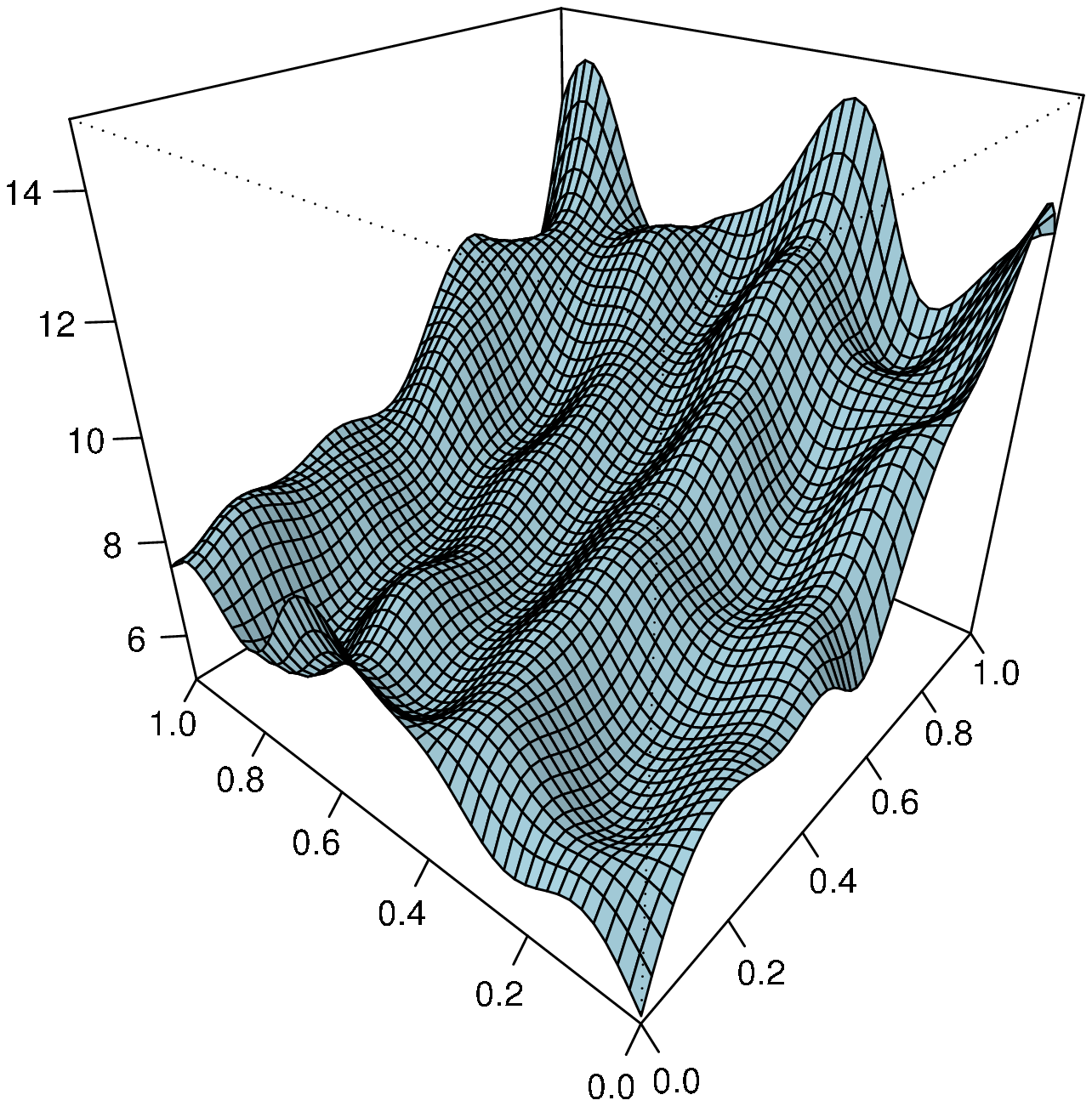} \\[1.cm]

 nonparametric additive SVM & &  semiparametric additive SVM \\[1.5cm]
\hspace{-1.2cm}
\includegraphics[width=4.6cm, angle=0]{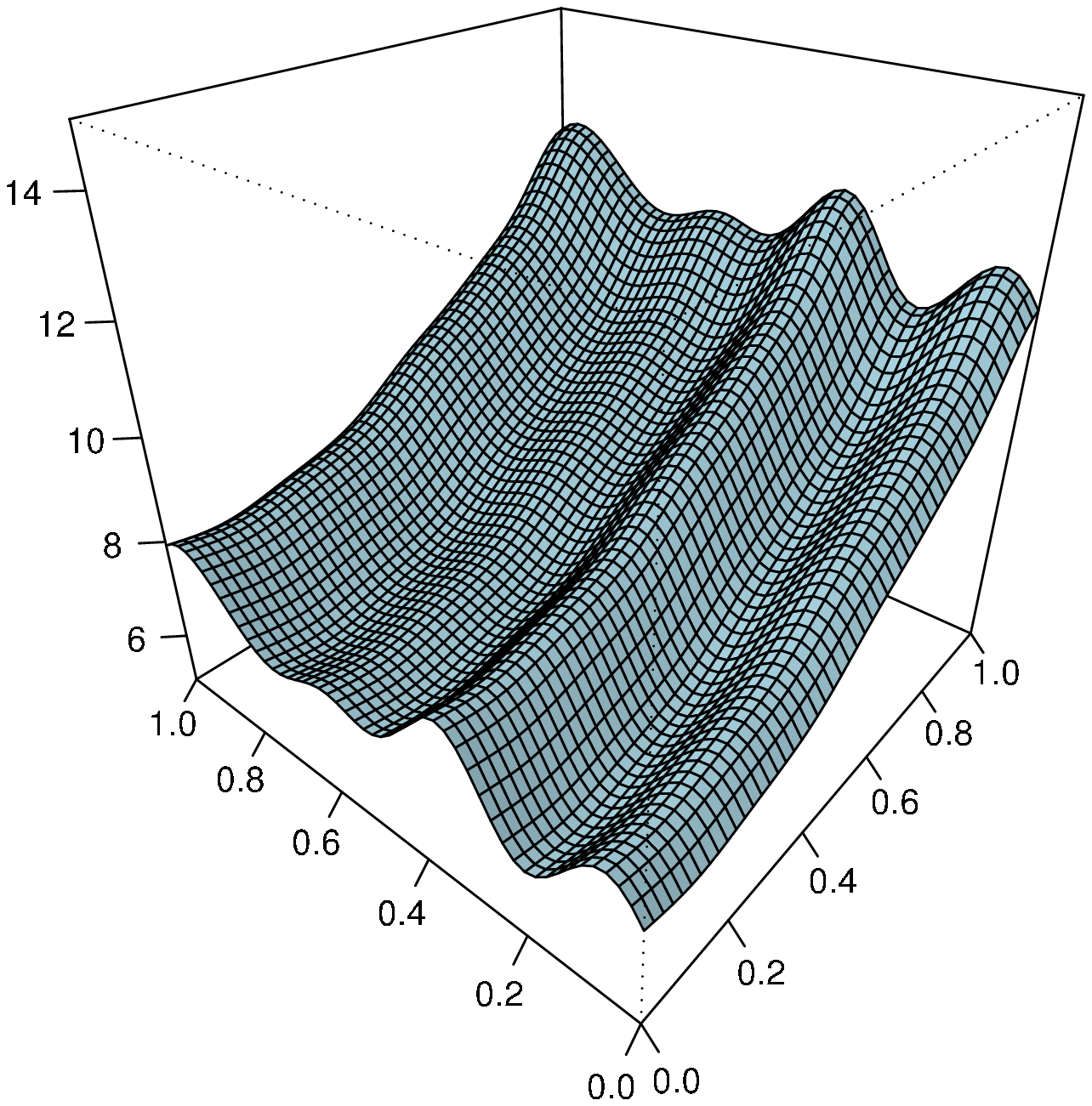} & &
\hspace{-2.0cm}
\includegraphics[width=4.6cm, angle=0]{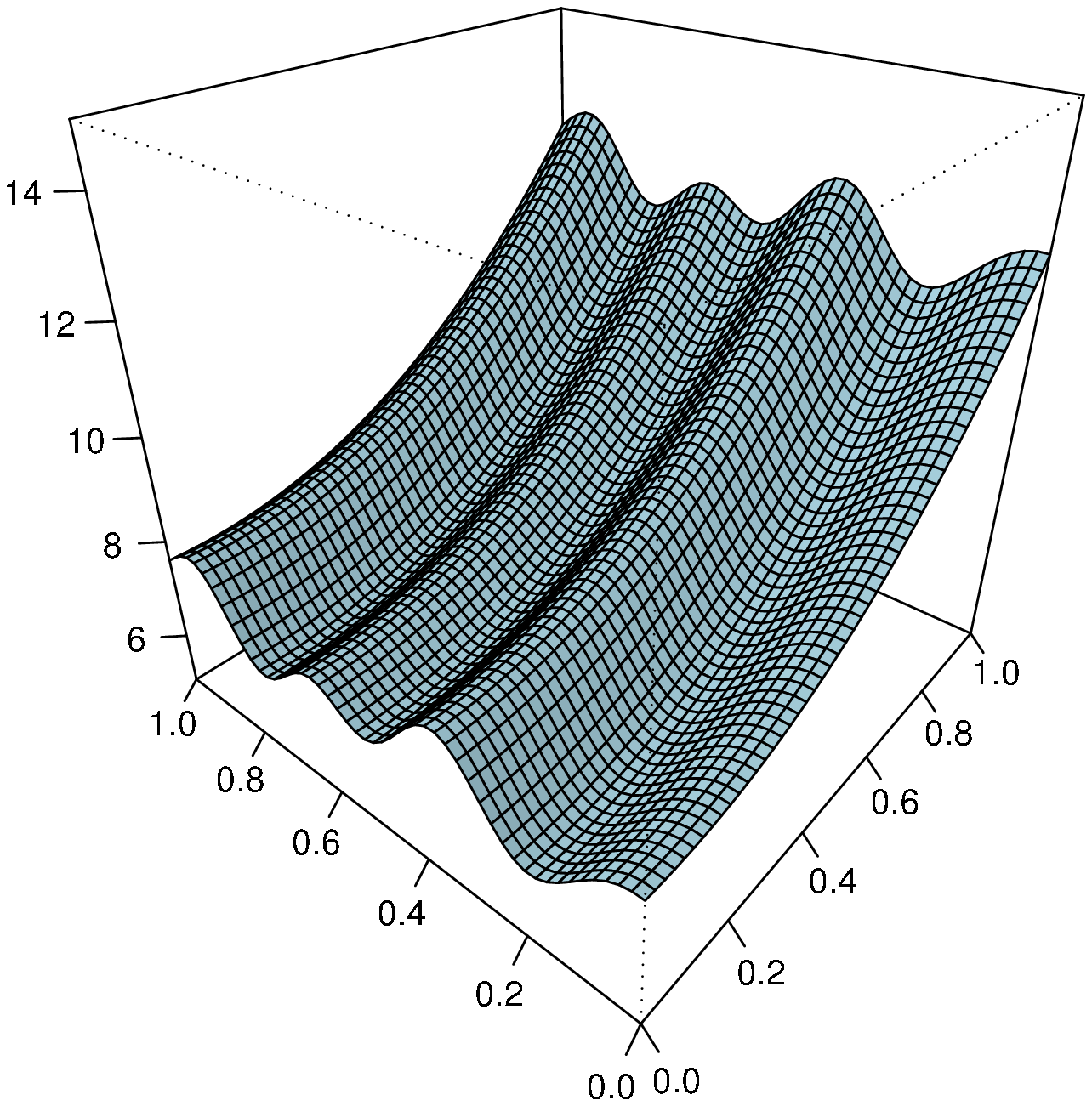} 
\end{tabular}
\end{figure}

\begin{figure}[t]
\caption{Quantile regression using SVMs and pinball loss function with $\tau=0.5$. Model: $Y|(x_1, x_2) \sim f_1(x_1)+f_2(x_2)+\mbox{Cauchy-errors}$, where $f_1(x_1):=7+5 x_1^2$ and $f_2(x_2)=\sin(5 x_2) \cos(17 x_2)$ and $x_1$ and $x_2$ are observations of independent and identically uniform distributed random variables on the interval $[0, 1]$. The regularization parameter is $\lb_n=0.05 n^{-0.45}$, and the kernel parameter of the Gaussian RBF kernel is $\gamma=2$. Upper left subplot: true function $f(x_1,x_2)=f_1(x_1)+f_2(x_2)$.
Upper right subplot: SVM fit based on GRBF kernel $k$ on $\cX=\R^2$.
Lower left subplot: SVM fit based on the sum of two 1-dimensional GRBF kernels.
Lower right subplot: SVM fit based on the sum of a 1-dimensional polynomial 
kernel on $\R$ and a 1-dimensional GRBF kernel. 
\label{FigureN500}\newline}
\begin{tabular}{ccc}
 \multicolumn{3}{c}{$\fbox{$n=500$}$} \\
 true function & $~~~~~~~~~~$ & nonparametric SVM \\[1.5cm]
\hspace{-1.2cm}  
\includegraphics[width=4.3cm,angle=0]{exact3d.eps} & &
\hspace{-2.0cm}
\includegraphics[height=4.6cm,angle=0]{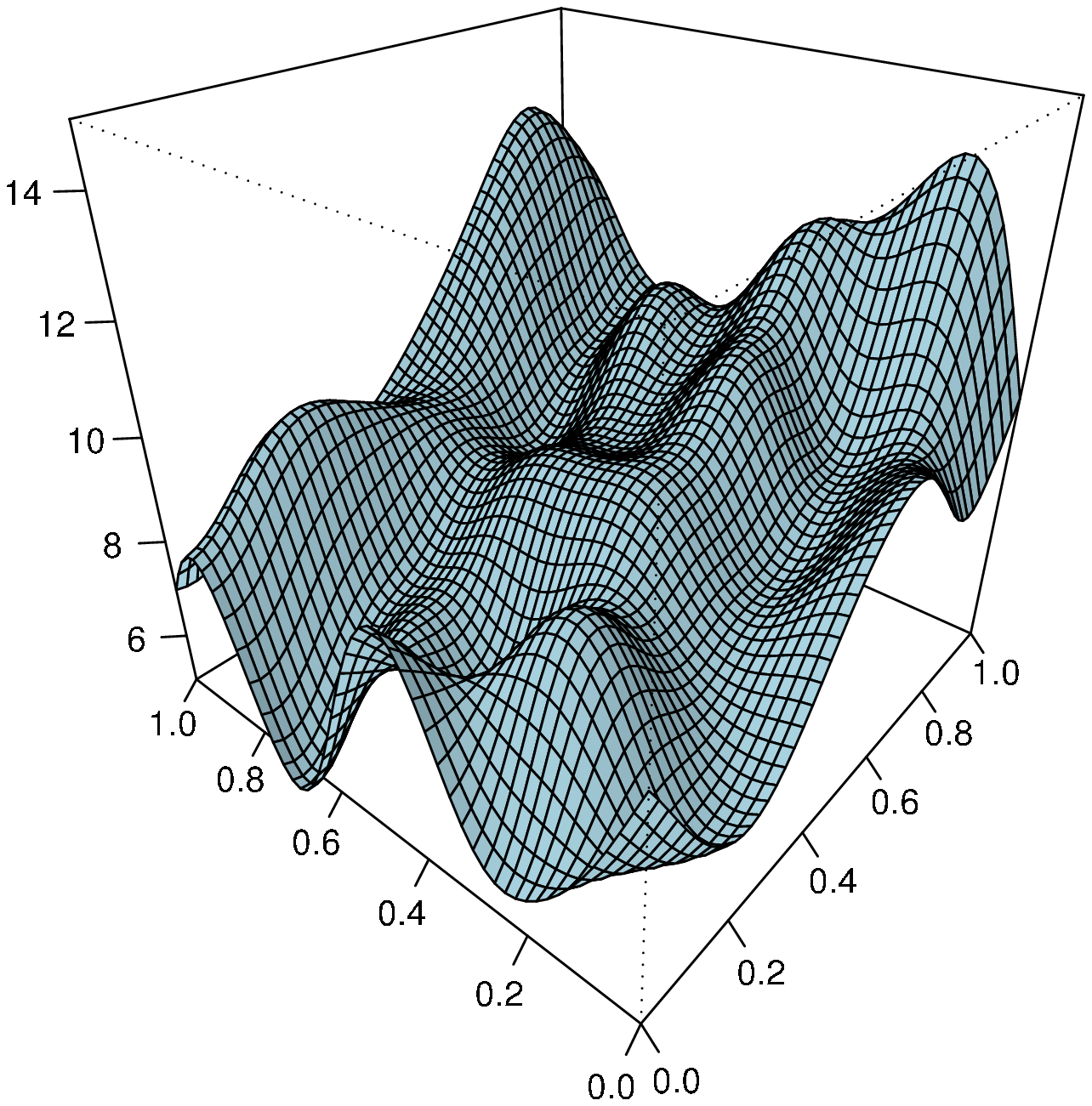} \\[1.cm]

 nonparametric additive SVM & &  semiparametric additive SVM \\[1.5cm]
\hspace{-1.2cm}
\includegraphics[width=4.6cm, angle=0]{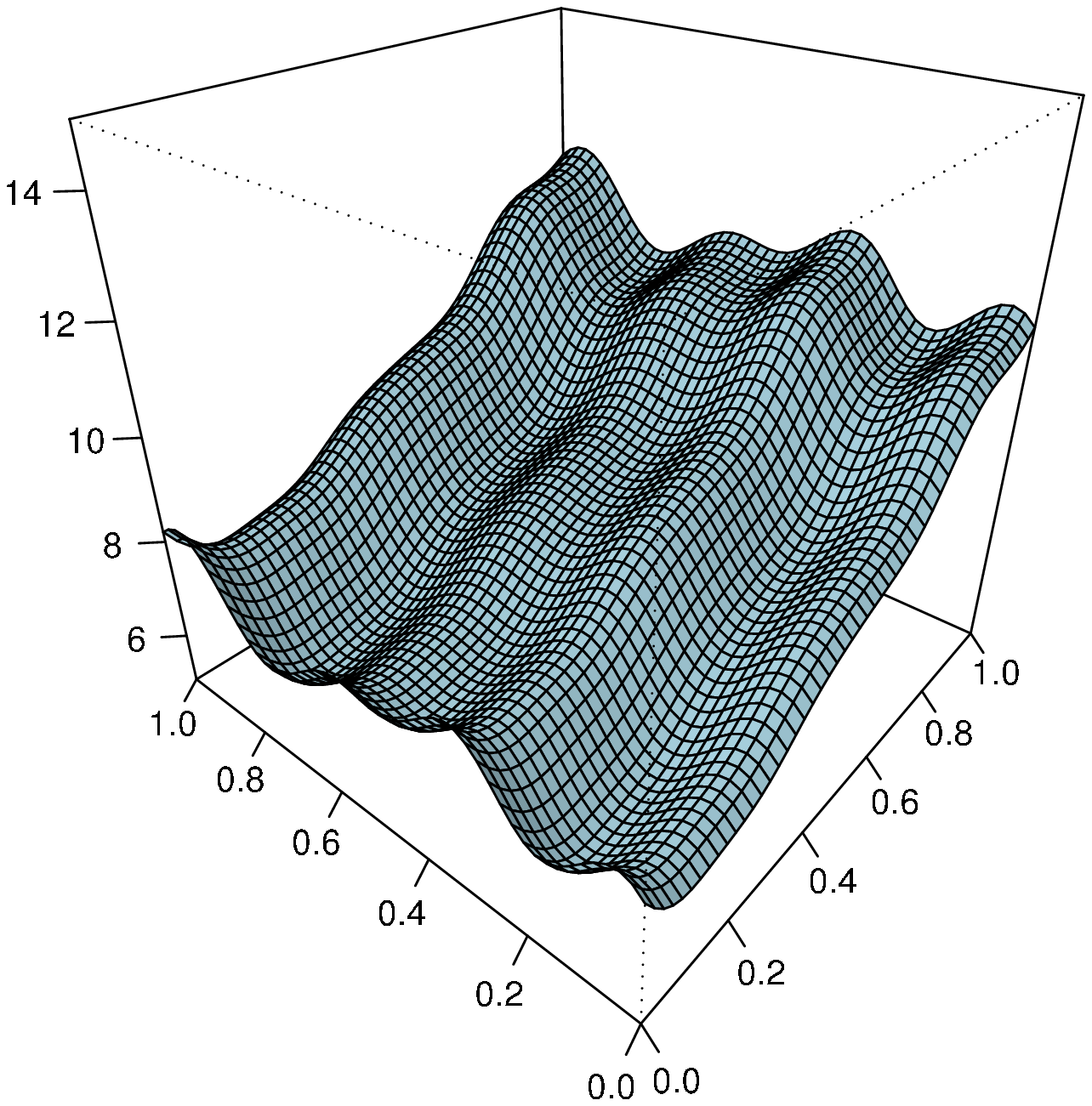} & &
\hspace{-2.0cm}
\includegraphics[width=4.6cm, angle=0]{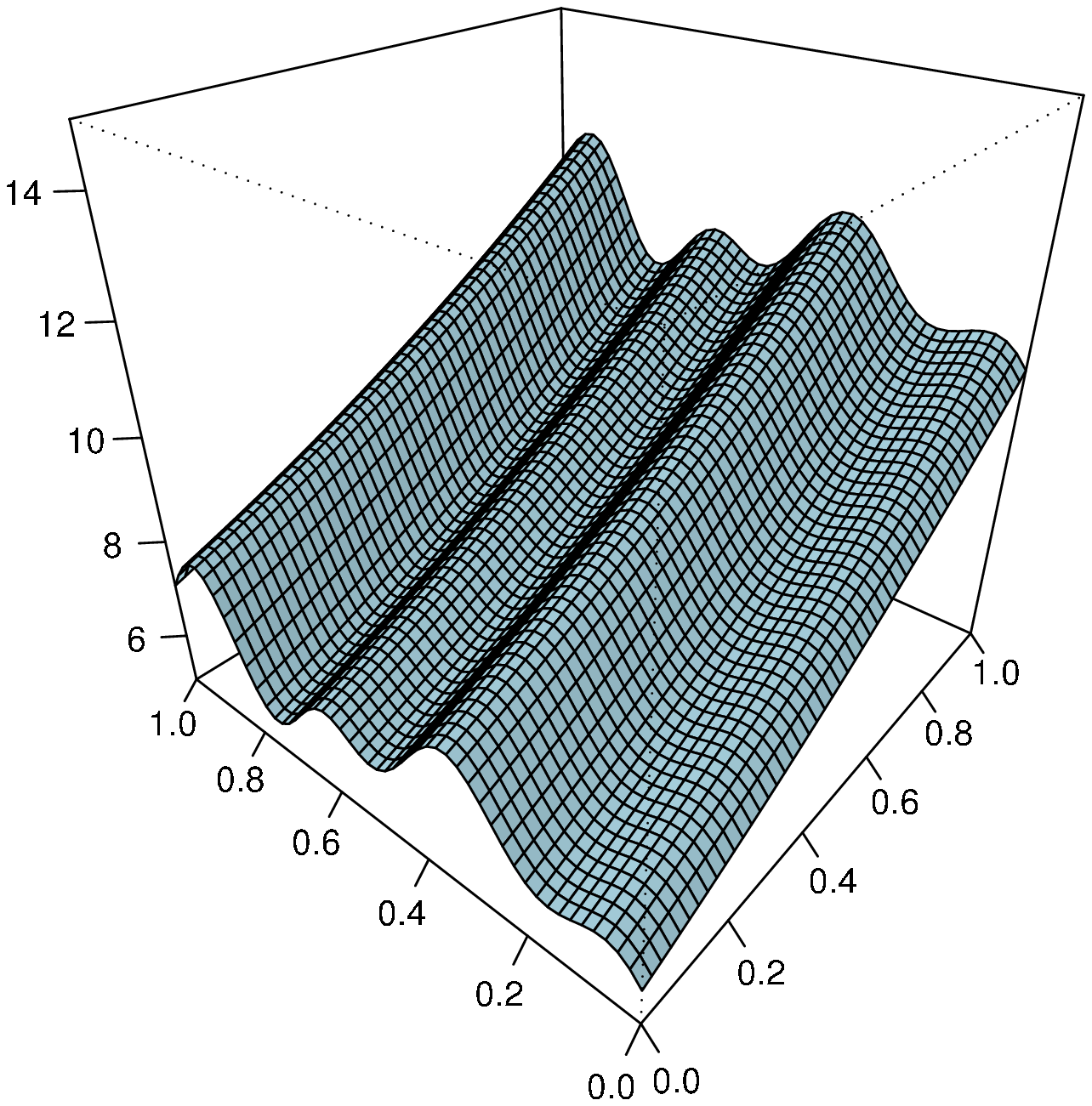} 
\end{tabular}
\end{figure}

From Figures \ref{FigureN10000} to \ref{FigureN500} we can draw the following conclusions for this special situation.
\bnum
\item If the additive model is valid, all three SVMs give comparable
      and reasonable results if the sample size $n$ is large enough even
      for Cauchy distributed error terms, see Figure \ref{FigureN10000}.
      This is in good agreement with the theoretical results derived in 
      Section \ref{sec:mainresult}.
\item If the sample size is small to moderate and if the assumed
      additive model is valid, then both SVMs based on kernels 
      especially designed for additive models show better results than
      the standard 2-dimensional GRBF kernel, see Figures \ref{FigureN2000}
      and Figure \ref{FigureN500}. 
\item The difference between the nonparametric additive SVM and 
      semiparametric additive SVM was somewhat surprisingly small for all three 
      sample sizes, although the true function had the very special structure which
      is in favour for the semiparametric additive SVM.
\enum

\subsection{Example: additive model using SVMs for rent standard}

Let us now consider a real-life example of the rent standard for dwellings in 
a large city in Germany.
Many German cities compose so-called rent standards to make a decision making
instrument available to tenants, landlords, renting advisory boards, and experts.
Such rent standards can in particular be used for the determination of the local
comparative rent, i.e. the net rent as a function of the dwelling size, year of construction of the house, geographical information etc. For the construction
of a rent standard, a representative random sample is drawn from all households and
questionnaires are used to determine the relevant information by trained interviewers.
\citet{FahrmeirKneibLang2007} described such a data set consisting of $n=3,\!082$ 
rent prizes in Munich, which is one of the largest cities in Germany. 
They fitted the following additive model
$$
   \mathrm{price} = f_1(\mathrm{size}) + f_2(\mathrm{year}) + 
           \beta_0 + \beta_1 \, \mathrm{region_1} + \beta_2 \,\mathrm{region_2} + 
           \mathrm{error},
$$
where the following variables were used:\\
\begin{tabular}{lll}
price &:& net rent price per square meter in DM [1 \officialeuro $~\approx$ 1.96 DM]\\
size &:& size in square meter of the dwelling [between 20 and 160]\\
year &:&  year [between 1918 and 1997] \\
$\mathrm{region_1}$ &:& good residential area [0=no, 1=yes] \\
$\mathrm{region_2}$ &:& best residential area [0=no, 1=yes].
\end{tabular}
Hence $\mathrm{region_1}$ and $\mathrm{region_2}$ are dummy variables with respect to
a standard residential area.
\citet{FahrmeirKneibLang2007} used a special spline method for estimating the functions $f_1$ and $f_2$.

For illustration purposes of SVMs with additive kernels investigated in the present paper, we used a nonparametric additive SVM for median regression. More precisely, 
we used the pinball loss function with $\tau=0.5$ and the kernel
$$
   k(x,x') = \sum_{j=1}^4 k_j(x_j,x_j'),  \quad x=(x_1,x_2,x_3,x_4)\in\R^4,~ x'=(x_1', x_2', x_3', x_4')\in\R^4,
$$
where\\
\begin{tabular}{ll}
$k_1: \R\to\R$ & Gaussian RBF kernel with $\gamma=1$ for size\\
$k_2: \R\to\R$ & Gaussian RBF kernel with $\gamma=1$ for year\\
$k_3: \R\to\R$ & dot kernel for $\mathrm{region_1}$\\
$k_4: \R\to\R$ & dot kernel for $\mathrm{region_2}$.
\end{tabular}

In analogy to the simulated examples given above, the regularizing parameter was again set to $\lambda_n=0.05 n^{-0.45}=0.00135$ such that our theoretical results are applicable.\footnote{Some numerical computations showed that the SVM results were fairly stable with respect to other choice of $\lambda_n$, e.g. $4\lambda_n$, for this particular data set and we will hence only show the results $\lambda_n=0.05 n^{-0.45}$.}

\begin{figure}[t]
\caption{Plot for the fitted additive model for the rent standard data set based on a nonparametric additive SVM for quantile regression, i.e., pinball loss function with $\tau=0.50$ (left) and $\tau=0.90$ (right). The surface gives the estimated median (left) or $90\%$ quantile (right) net rent price of one square meter depending on the size of the dwelling and the year of the construction for a standard residential area, i.e., $\mathrm{region_1}=\mathrm{region_2}=0$. 
\label{fig:mietspiegeltau5090}}
\bc
\begin{tabular}{cc}
$\tau=0.5$ & $\tau=0.9$ \\
\includegraphics[width=6.4cm,angle=-90]{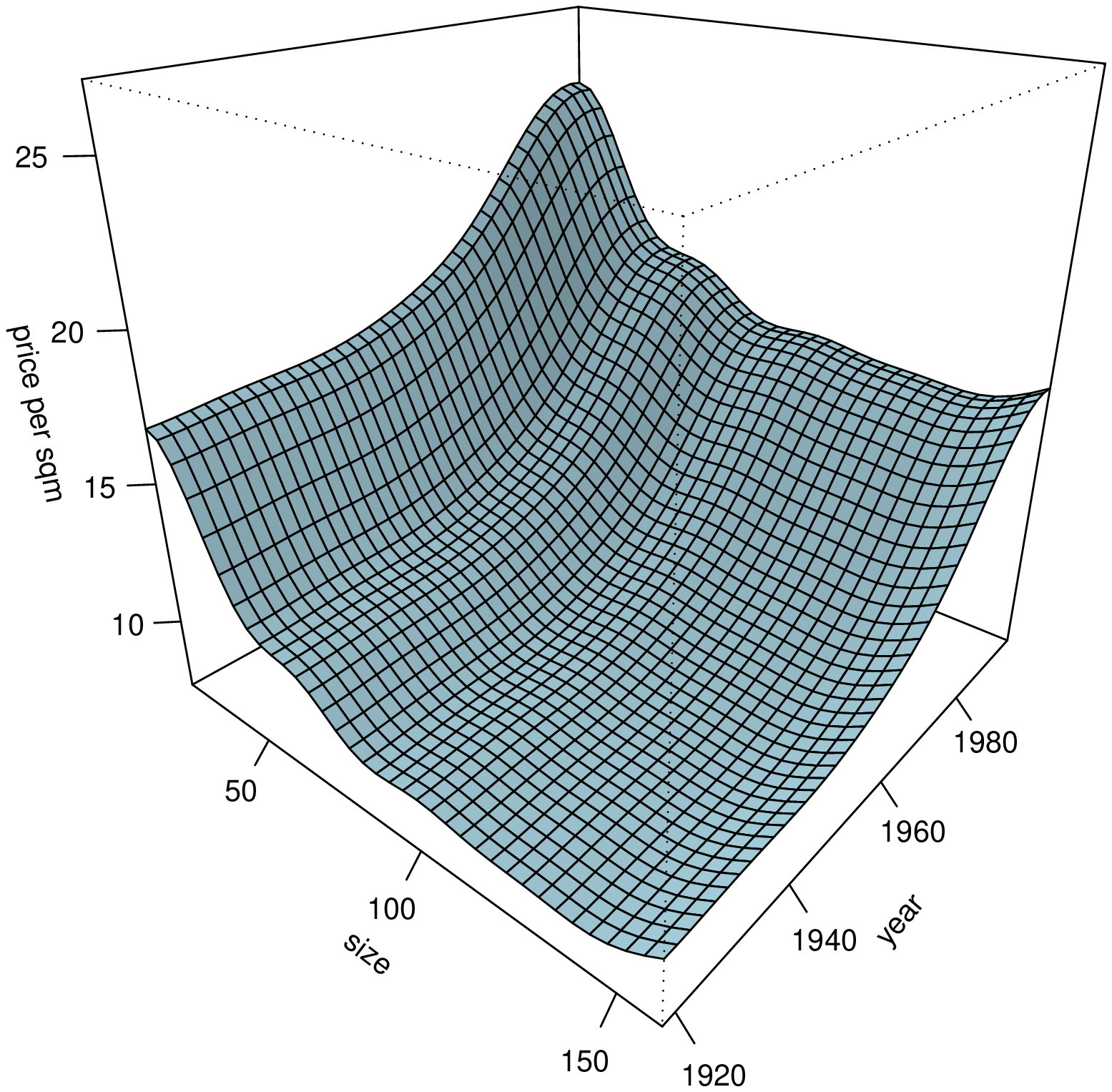} &
\includegraphics[width=6.8cm,angle=-90]{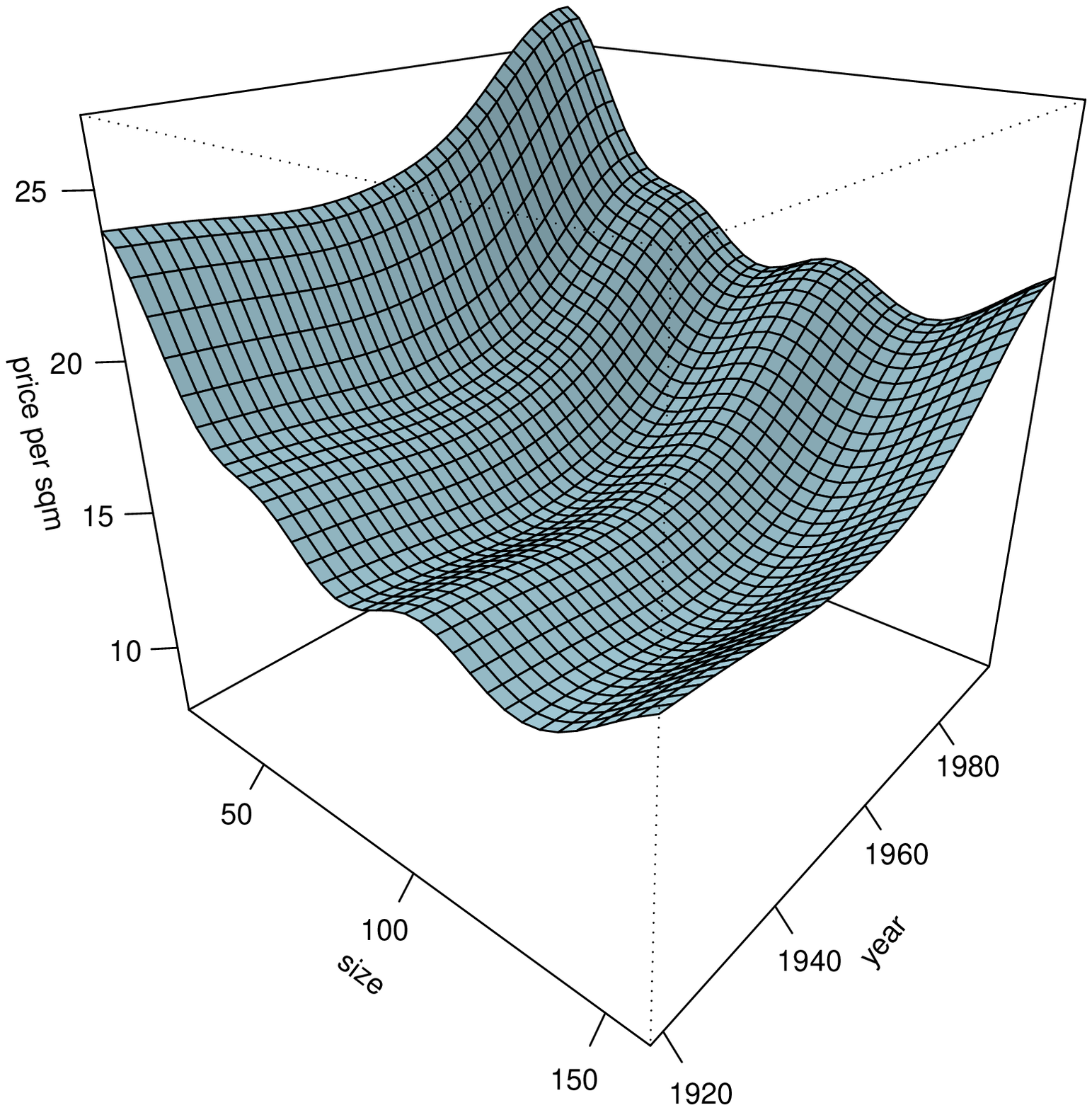}
\end{tabular}
\ec
\end{figure}

The left plot in Figure \ref{fig:mietspiegeltau5090} shows the estimated median net rent price of one square meter depending on the size of the dwelling and the year of the construction for a dwelling in a standard residential area.
The plot shows that the median of the net rent prices per square meter is fairly stable for construction years up to 1960, but a more or less linear increase is visible for newer buildings.
The plot also shows that the median of the net rent prices per square meter is especially high for dwellings of size less than 80 square meter, that the price is nearly constant for sizes between 80 and 140 square meter, and then
a slight increase of the square meter prize seems to occur for even larger
dwellings. 
The slope parameters were estimated by
$\hat{\beta}_1= 1.38$ for good residential area ($\mathrm{region_1}=1$) and 
$\hat{\beta}_1= 3.46$ for best residential area ($\mathrm{region_2}=1$). Hence, we obtain apart from these level shifts the same surfaces for dwellings located in good or in best residential areas.
We would like to mention that we used this real-life example just for illustration purposes, but nevertheless our results are in good agreement with the more detailed statistical analysis  of this data set made by \citet{FahrmeirKneibLang2007} who used different statistical techniques.

From an applied point of view, one may also be interested in the $10\%$ highest net rent prices depending on the four explanatory variables.
We therefore repeated our computations using the same kernel but instead of $\tau=0.50$ for median regression we used $\tau=0.90$ to obtain estimates for the $90\%$ quantiles
of the net rent prizes depending on the four explanatory variables. 
The right plot in Figure \ref{fig:mietspiegeltau5090} shows the estimated $90\%$ quantile net rent prices of one square meter depending on the size of the dwelling and the year of the construction for a dwelling in a standard residential area. 
The slope parameters were estimated by
$\hat{\beta}_1= 1.59$ for good residential area ($\mathrm{region_1}=1$) and 
$\hat{\beta}_1= 4.24$ for best residential area ($\mathrm{region_2}=1$).
The shape of the surface is quite similar to the shape of the surface in the previous plot for the estimated median net rent prices.
However, the plot may give an indication for a moderate peak for the  
$90\%$ quantile net rent prices for dwellings of size 100 square meter.

\section{Discussion} \label{sec:discussion}
  
Support vector machines belong to the class of modern statistical machine learning methods based on kernels. The success of SVMs is partly based on
on the kernel trick which makes SVMs usuable even for abstract input spaces, their universal consistency, their statistical robustness with respect to small model violations, and on the existence of fast numerical algorithms. 
During the last decade there has been considerable research on these three 
topics. To obtain universal consistency one needs a sufficiently large
reproducing kernel Hilbert space $H$, such that many interesting 
SVMs are based on Hilbert spaces with infinite dimension.
Due to the no-free-lunch theorem \citep{Devroye82a}, there exists in general 
no uniform rate of convergence of SVMs on the set of all probability measures.

Although such a nonparametric approach is often the best choice in practice due to the lack of prior knowledge on the unknown probability measure $\P$, a semiparametric approach or an additive model \citep{FriedmanStuetzle1981,HastieTibshirani1990} can also be valuable for at least two reasons:
\emph{(i)} In some applications some weak knowledge on $\P$ or on the
unknown function $f$ to be estimated, say the conditional quantile curve, is known, e.g. $f$ is known to be bounded or at least integrable.
\emph{(ii)} Due to practical reasons, we may be interested only in functions $f$ which offer a nice interpretation although there might be a measurable function with a smaller risk, because an interpretable prediction function can be crucial in some applications. An important class of statistical models whose predictions are relatively easily to interpret are additive models.

Therefore, support vector machines for additive models were treated in this paper and some results on their consistency and statistical robustness properties were derived. 

Some simple numerical examples showed that SVMs based on kernels especially designed for an additive model can yield better predictions than the standard SVM based on the classical Gaussian RBF kernel, if an additive model is indeed valid. 

It may be worthwhile to investigate the rates of convergence of SVMs based on kernels designed for additive models with SVMs based on standard kernels, because our simple numerical examples seem to indicate that there might be some gain with respect to the rate of convergence.  However, this is beyond the scope of this paper.

We would like to mention the well-known fact that not only the sum of $s$ kernels is a kernel but also the product of $s$ kernels is a kernel.
Hence it seems to be possible to derive similar results than those given here
for additive models also for multiplicative models.

Finally, we would like to mention that there are of course many other
statistical estimation techniques for additive models, e.g. 
splines and boosting, but a comparision of these methods with SVMs based on additive kernels is beyond the scope of this paper.

\appendix
\section*{Appendix: Proofs} \label{sec:appendix}
\setcounter{section}{1}
\renewcommand{\Theorem}{A.Definition}
\setcounter{equation}{0}

\begin{proofof}{Proof of Theorem %
                \ref{theorem-sum-of-kernels-on-different-domains}%
               }
  First fix any $j\in\{1,\dots,s\}$ and define the mapping
  $\tilde{k}_j:\cX\times\cX\rightarrow\R$ via
  $$\tilde{k}_j\big((x_1,\dots,x_s),(x_1^\prime,\dots,x_s^\prime)\big)
    \;\;=\;\;k_j(x_j,x_j^\prime)
  $$
  for every 
  $(x_1,\dots,x_s)\in\cX$ and $(x_1^\prime,\dots,x_s^\prime)\,\in\,\cX$.
  Accordingly, for every $f_j\in H_j$, define 
  $\tilde{f}_j:\cX\rightarrow\R$ via
  $$\tilde{f}_j(x_1,\dots,x_s)\;=\;f_j(x_j)
    \qquad\quad\forall\,(x_1,\dots,x_s)\in\cX\;.
  $$
  Then, it is easy to see
  that
  $$\tilde{H}_j\;=\;
    \big\{\tilde{f}_j:\cX\rightarrow\R\;:\;\;
          f_j\in H_j
    \big\}
  $$
  is a Hilbert space with inner product and norm given by
  \begin{eqnarray}\label{theorem-sum-of-kernels-on-different-domains-p1}
    \langle \tilde{f}_j, \tilde{h}_j\rangle_{\tilde{H}_j}
    \;=\;\langle f_j,h_j\rangle_{H_j}
    \qquad\text{and}\qquad
    \big\|\tilde{f}_j\big\|_{\tilde{H}_j}
    \;=\;\big\| f_j\big\|_{H_j}
  \end{eqnarray}
  for every $f_j\in H_j$ and $g_j\in H_j$. Hence,
  for every $x=(x_1,\dots,x_s)\in\cX$, we get 
  $\tilde{k}_j(\cdot,x)\in\tilde{H}_j$ 
  and
  $$\tilde{f}_j(x)=f_j(x_j)
    =\langle f_j,k_j(\cdot,x_j)\rangle_{H_j}
    =\langle \tilde{f}_j, \tilde{k}_j(\cdot,x)\rangle_{\tilde{H}_j}
    \qquad\forall\,f_j\in H_j
  $$
  where the last equality follows from 
  (\ref{theorem-sum-of-kernels-on-different-domains-p1}) and
  the definition of $\tilde{k}_j$. That is, $\tilde{k}_j$
  is a reproducing kernel and $\tilde{H}_j$ is its 
  RKHS. 

  Next, it follows from \citep[\S\,4.1]{berlinet2004}
  that $k=\tilde{k}_1+\dots+\tilde{k}_s$ is a reproducing kernel
  on $\cX$
  with RKHS $H=\tilde{H}_1+\dots+\tilde{H}_s$ and norm
  \begin{eqnarray}\label{theorem-sum-of-kernels-on-different-domains-p2}
    \big\|f\big\|_H^2
    &=&
    \min_{f=\tilde{f}_1+\dots+\tilde{f}_s \atop 
          \tilde{f}_1\in\tilde{H}_1,\dots,\tilde{f}_s\in\tilde{H}_s %
         }\!
    \big\|\tilde{f}_1\big\|_{\tilde{H}_1}^2
    +\dots+\big\|\tilde{f}_s\big\|_{\tilde{H}_s}^2\;=\nonumber\\
    &\stackrel{(\ref{theorem-sum-of-kernels-on-different-domains-p1})}{=}&
    \min_{f=\tilde{f}_1+\dots+\tilde{f}_s \atop 
          f_1\in H_1,\dots, f_s\in H_s %
         }\!
    \big\|f_1\big\|_{H_1}^2+\dots+\big\|f_s\big\|_{H_s}^2
  \end{eqnarray}
  Using the reduced notation 
  $f_1+\dots+f_s$ instead of
  $\tilde{f}_1+\dots+\tilde{f}_s$, inequality 
  (\ref{theorem-sum-of-kernels-on-different-domains-1}) follows.
  \qedr
\end{proofof}

In order to prove Theorem \ref{theorem-consistency}, the following 
proposition is needed. It provides conditions on
$H_j$ and $\cF_j$ under which the minimal risk over
$H=H_1+\dots+H_s$ is equal to the minimal risk
over the larger $\cF=\cF_1+\dots+\cF_s$.
\begin{Proposition}\label{prop-density-risks}
  Let the main assumptions (p.\ \pageref{assumptions}) be valid. Let
  $\P\in\mathcal{M}_1(\cXY)$ such that
  $$H_j\;\subset\;\cF_j\;\subset\;\mathcal{L}_1(\P_{\cX_j})
    \qquad\forall\,j\in\{1,\dots,s\}
  $$
  and $H_j$ is dense in $\cF_j$ with respect to
  $\|\cdot\|_{L_1(\P_{\cX_j})}$. Then,
  \begin{eqnarray}\label{prop-density-risks-1}
    \mathcal{R}_{\Ls,\P,H}^{\ast}\;:=\;
    \inf_{f\in H}\mathcal{R}_{\Ls,\P}(f)
    \;=\;\mathcal{R}_{\Ls,\P,\cF}^{\ast}\;.
  \end{eqnarray} 
\end{Proposition}
\begin{proofof}{Proof of Proposition \ref{prop-density-risks}}
  According to the definitions, it only remains to prove
  $\mathcal{R}_{\Ls,\P,H}^{\ast}\leq\mathcal{R}_{\Ls,\P,\cF}^{\ast}$\,.
  To this end,
  take any $f\in\cF$ and any $\varepsilon>0$. Then, by assumption
  there are functions
  $$f_j\in\cF_j\,,\qquad j\in\{1,\dots,n\},
  $$
  such that $f=f_1+\dots+f_s$ and, for every $j\in\{1,\dots,s\}$,
  there is an $h_j\in H_j$ such that
  \begin{eqnarray}\label{prop-density-risks-p1}
    \big\|h_j-f_j\big\|_{L_1(\P_{\cX_j})}\;<\;
    \frac{\varepsilon}{s\cdot|L|_1}\;.
  \end{eqnarray}
  Hence, for $h=h_1+\dots+h_s\,\in\,H$,
  \begin{eqnarray*}
    \lefteqn{
      \big|\RP{\Ls}{h}-\RP{\Ls}{f}\big|
      \;\leq\;\int\big|L(x,y,h(x))-L(x,y,f(x)\big|\,\P\big(d(x,y)\big)\;
    }\\ 
    &\leq&|L|_1\int\big|h(x)-f(x)\big|\,\P_{\cX}(dx)
          \;\leq\; |L|_1\sum_{j=1}^{s}
                   \int\big|h_j(x_j)-f_j(x_j)\big|\,\P_{\cX_j}(dx_j)\\
    &<&\varepsilon
  \end{eqnarray*}
  \qedr
\end{proofof}

\begin{proofof}{Proof of Theorem \ref{theorem-consistency}}
  To avoid handling too many constants, let us 
  assume $\inorm{k}=1$. 
  According to (\ref{wk1a}),
  this implies $\inorm{f} \le \hnorm{f}$ for all $f\in \cH$. 
  Now we use the Lip\-schitz continuity
  of $L$ to obtain, 
  for all $g\in \cH$,  
  \begin{eqnarray}\label{svmregsec:4:cons-h1}
    \lefteqn{
      \bigl| \RP{\Ls}{f_{L,\P,\lb_n}}- \RP{\Ls}{g}  \bigr|\;\leq
    }\nonumber\\
    &\leq&\int\big|L(x,y,f_{L,\P,\lb_n}(x))
                   -L(x,y,g(x))\big|\,\P\big(d(x,y)\big)\;\nonumber\\
    &\leq& |L|_1\int\big|f_{L,\P,\lb_n}(x)-g(x)\big|\,\P_{\cX}(dx)
          \;\leq\;|L|_1
                  \int\big\|f_{L,\P,\lb_n}-g\big\|_{\infty}\,\P_{\cX}(dx)
          \quad\nonumber\\
    &\leq&|L|_1 \, \hnorm{f_{\Ls,\P,\lb_n}-g}.  
  \end{eqnarray} 
  Let $\Phi$ denote the canonical feature map which corresponds to
  the kernel $k$.
  According to \citet[Theorem 7]{ChristmannVanMessemSteinwart2009}, 
  for every $n\in\N$, there is
  a bounded, measurable function $h_n:\cXY\rightarrow\R$
  such that
  \begin{eqnarray}\label{theorem-consistency-p1}
    \inorm{h_n} \le |L|_1
  \end{eqnarray}
  and, for every $\Q\in\mathcal{M}_1(\cXY)$,
  \begin{eqnarray}\label{theorem-consistency-p2}
    \hnorm{f_{L,\P,\lb_n}-f_{L,\Q,\lb_n}}
    \le
    \lb_n^{-1} \, \hnorm{\Ex_\P h_n\Phi - \Ex_\Q h_n\Phi}
    \;.
  \end{eqnarray}
  Fix any $\ve\in(0,1)$ and
  define
  \begin{equation}\label{svmregsec:4:cons-h2}
    B_n\;:=\;
    \left\{D_n\in(\cXY)^n\;:\;\;
          \hnorm{ \Ex_\P h_n \Phi - \Ex_{\D_n} h_n \Phi} 
          \leq  \frac{\lb_n \ve}{|L|_1}
    \right\}
  \end{equation}
  where $\D_n$ denotes the empirical distribution
  of the data set $D_n$.
  Then, {(\ref{svmregsec:4:cons-h1})}, (\ref{theorem-consistency-p2}) and 
  (\ref{svmregsec:4:cons-h2}) yield
  \begin{equation}\label{svmregsec:4:cons-h3}
    \bigl| \RP{\Ls}{f_{L,\P,\lb_n}}- \RP{\Ls}{f_{L,\D_n,\lb_n}} \bigr|
    \;\leq\; \ve 
    \qquad\forall\,D_n\in B_n\; .
  \end{equation}
  Now let us turn over to the probability $\P^n(B_n)$. By use of
  Hoeffding's inequality, we will show that
  \begin{eqnarray}\label{theorem-consistency-p3}
    \lim_{n\rightarrow\infty}\P^n(B_n)\;=\;1
    \;.
  \end{eqnarray}
  To this end, we first observe that
  $\lb_n n^{1/2}\to \infty$ implies that $\lb_n \ve \ge n^{-1/2}$
  for all sufficiently large $n\in\N$. Moreover, 
  (\ref{theorem-consistency-p1}) and
  our assumption $\inorm{k}=1$ yield
  $\inorm{h_n \Phi} \le|L|_1$. 
  Define
  $$a_n\;:=\;|L|_1^{-1} \ve \lb_n
    \qquad\text{and}\qquad 
    \xi_n\;:=\;\frac{3}{8} \, \frac{  |L|_1^{-2} \ve^2\lb_n^2 n }
                       {  |L|_1^{-1} \ve \lb_n+3  }
         \;=\;\frac{3}{8} \, \frac{  a_n^2 n }
                       {  a_n+3  }                       
  $$
  and note that, for sufficiently large $n$,
  \begin{eqnarray}\label{theorem-consistency-p5}
    \frac{\sqrt{2\xi_n}+1}{\sqrt{n}}+\frac{4\xi_n}{3n}
    &=&\frac{a_n}{2}\cdot\frac{\sqrt{3}}{\sqrt{a_n+3}}+\frac{1}{\sqrt{n}}
           +\frac{a_n}{2}\cdot\frac{a_n}{a_n+3}
        \nonumber \\
    &<&\frac{a_n}{2}+\frac{1}{\sqrt{n}}+\frac{a_n}{2}\cdot\frac{1}{3}
          \;<\;a_n\;=\;|L|_1^{-1} \ve \lb_n\;.\qquad
  \end{eqnarray}
  Consequently, Hoeffding's inequality in Hilbert spaces 
  (see \citet[Corollary 6.15]{SteinwartChristmann2008a}) yields for $B=1$ 
  the bound 
  \begin{eqnarray*}
    \lefteqn{
      \P^n(B_n)
      \;=\;\P^n \left( \left\{D \in (\cXY)^n:\, 
                                \hnorm{\Ex_\P h_n\Phi -\Ex_{\D} h_n\Phi} 
                              \le \frac{\lb_n \ve}{|L|_1} 
                       \right\}
                \right) 
       \;\geq
   }\\
   &\stackrel{(\ref{theorem-consistency-p5})}{\geq}&
         \P^n \left( \left\{D \in (\cXY)^n:\, 
                            \hnorm{\Ex_\P h_n\Phi - \Ex_{\D} h_n\Phi} \le 
                            \frac{\sqrt{2\xi}+1}{\sqrt{n}}
                            +\frac{4\xi}{3n}
                     \right\} 
              \right)  \\
   &\geq& 1 - 
          \exp\Bigl(-\frac{3}{8}\cdot
                     \frac{\ve^2\lb_n^2 n/|L|_1^2}{\ve \lb_n/|L|_1+3} 
              \Bigr) 
       \;=\;1 - 
         \exp\Bigl(- \frac{3}{8} \cdot 
                    \frac{\ve^2 \lb_n^2 n}{(\ve \lb_n + 3|L|_1)|L|_1} 
             \Bigr)         
  \end{eqnarray*}
  for all sufficiently large values of $n$. Now 
  (\ref{theorem-consistency-p3}) follows from
  $\lb_n\to 0$ and $\lb_n n^{1/2} \to \infty$. 
  According to (\ref{svmregsec:4:cons-h3}) and 
  (\ref{theorem-consistency-p3}),
  $$\RP{\Ls}{f_{L,\P,\lb_n}}- \RP{\Ls}{\fDDnln}\;\longrightarrow\;0
    \qquad\quad(n\rightarrow\infty)
  $$
  in probability.
  Note that,
  \begin{eqnarray}\label{theorem-consistency-p4}
    \lefteqn{
      \Big|\RP{\Ls}{\fDDnln}-\mathcal{R}_{\Ls,\P,\cF}^{\ast}\Big|\;\leq
    } \nonumber \\
    &\leq&\!\!
          \Big|\RP{\Ls}{\fDDnln}-\mathcal{R}_{\Ls,\P,H}^{\ast}\Big|
          +\Big|\mathcal{R}_{\Ls,\P,H}^{\ast}-\mathcal{R}_{\Ls,\P,\cF}^{\ast}
           \Big| \nonumber  \\
    &\stackrel{(\ref{prop-density-risks-1})}{\leq}&\!\!
          \Big|\RP{\Ls}{\fDDnln}-\RP{\Ls}{f_{L,\P,\lb_n}}\Big|
          +\Big|\RP{\Ls}{f_{L,\P,\lb_n}}-\mathcal{R}_{\Ls,\P,H}^{\ast}\Big|
          \qquad\quad
  \end{eqnarray}
  As shown above, the first term in (\ref{theorem-consistency-p4})
  converges in probability to 0. Therefore, it only remains to
  prove that the second term converges to 0.
  To this end, define, for every $f\in H$, the
  affine linear function
  $$A_{f}^\ast\;:\;\;\R\;\rightarrow\;\R\,,\qquad
    \lambda\;\mapsto\;\RP{\Ls}{f}+\lambda\|f\|_H^2
                      -\mathcal{R}_{\Ls,\P,H}^{\ast}\;.
  $$
  Then, a continuity result for the pointwise infimum
  of a family of affine functions 
  (see e.g.\ \citep[A.6.4]{SteinwartChristmann2008a}) yields
  $$\lim_{n\rightarrow\infty}\inf_{f\in H}A_{f}^\ast(\lambda_n)
    \;=\;\inf_{f\in H}A_{f}^\ast(0)\;.
  $$
  However, according to the definitions,
  $$\inf_{f\in H}A_{f}^\ast(\lambda_n)
    \;=\;\RP{\Ls}{f_{L,\P,\lb_n}}+\lambda_n\|f_{L,\P,\lb_n}\|_H^2
         -\mathcal{R}_{\Ls,\P,H}^{\ast}
    \qquad\forall\,n\in\N
  $$
  and
  ${\displaystyle
    \,\inf_{f\in H}A_{f}^\ast(0)\,=\,0\,.
   }
  $
  Hence,
  \begin{eqnarray*}
    0
    &\leq&\limsup_{n\rightarrow\infty}\,
           \big(\RP{\Ls}{f_{L,\P,\lb_n}}-\mathcal{R}_{\Ls,\P,H}^{\ast}
           \big)\;\leq\\
    &\leq&\limsup_{n\rightarrow\infty}
           \Big(\inf_{f\in H}A_{f}^\ast(\lambda_n)
                -\inf_{f\in H}A_{f}^\ast(0)
           \Big)
          \;=\;0
  \end{eqnarray*}
\qedr
\end{proofof}

\begin{proofof}{Proof of Theorem \ref{theorem-quantile-regression-consistency}}
  Since the quantile function $f^\ast_{\tau,\P}$ attains the
  minimal risk $\mathcal{R}_{\Ls,\P}^{\ast}$ for the
  pinball loss $L=L_\tau$ \citep[\S\,1.3]{Koenker2005}, 
  the assumption $f^\ast_{\tau,\P}\in\cF$ implies
  $\mathcal{R}_{\Ls,\P,F}^{\ast}=\mathcal{R}_{\Ls,\P}^{\ast}$.
  Hence, an application of Theorem \ref{theorem-consistency} yields
  \begin{eqnarray}\label{theorem-quantile-regression-consistency-p1}
    \RP{\Ls}{\fDDnln}\;\longrightarrow\;\mathcal{R}_{\Ls,\P}^{\ast}
    \qquad\quad(n\rightarrow\infty)
  \end{eqnarray}
  in probability. It is shown in
  \cite[Corollary 31]{ChristmannVanMessemSteinwart2009}
  that, for all sequences $(f_n)_{n\in\N}$ of measurable
  functions $f_n:\cX\rightarrow\R$, 
  $$\RP{\Ls}{f_n}\;\longrightarrow\;\mathcal{R}_{\Ls,\P}^{\ast}
    \qquad\text{implies}\qquad
    d_0\big(f_n,f^\ast_{\tau,\P}\big)\;\longrightarrow\;0\;.
  $$
  This proves Theorem \ref{theorem-quantile-regression-consistency}
  in the following way:
  According to the characterization
  of convergence in probability by means of
  almost surely convergent subsequences 
  \citep[Theorem 9.2.1]{Dudley2002},
  it follows from (\ref{theorem-quantile-regression-consistency-p1})
  that, for every subsequence of
  $\RP{\Ls}{\fDDnln}$, $n\in\N$, there is a 
  further subsequence which converges almost surely to
  $\mathcal{R}_{\Ls,\P}^{\ast}$. Hence, according to
  the cited result 
  \citep[Corollary 31]{ChristmannVanMessemSteinwart2009},
  for every subsequence of
  $$d_0\big(\fDDnln,f^\ast_{\tau,\P}\big)\,,\quad
    n\in\N\,,
  $$
  there is a 
  further subsequence which converges almost surely to
  $0$. That is, 
  $d_0\big(\fDDnln,f^\ast_{\tau,\P}\big)\rightarrow0$
  in probability.
\qedr
\end{proofof}

  
\bibliographystyle{plainnat}
\bibliography{christmann,hable}

\end{document}